\documentclass[lettersize,journal]{IEEEtran}
\usepackage{amsmath,amsfonts}
\usepackage{algorithmic}
\usepackage{algorithm}
\usepackage{array}
\usepackage[caption=false,font=normalsize,labelfont=sf,textfont=sf]{subfig}
\usepackage{textcomp}
\usepackage{amsmath}     
\usepackage{stfloats}
\usepackage{bibentry}
\usepackage{url}
\usepackage{verbatim}
\usepackage{graphicx}
\usepackage{cite}
\usepackage{multirow}
\usepackage{balance}
\usepackage{subfloat}
\usepackage{threeparttable}
\usepackage{makecell}
\usepackage{stfloats}
\usepackage{amssymb}    
\usepackage{booktabs}
\usepackage{array}
\usepackage{siunitx}
\usepackage{xspace}
\usepackage{ragged2e}
\usepackage{pifont}
\usepackage{gensymb}

\def\BibTeX{{\rm B\kern-.05em{\sc i\kern-.025em b}\kern-.08em
		T\kern-.1667em\lower.7ex\hbox{E}\kern-.125emX}}
\usepackage{balance}
\begin{document}
	\title{Environment-Aware Channel Prediction for Vehicular Communications: A Multimodal Visual Feature Fusion Framework}    
	\author{Xuejian Zhang, ~\IEEEmembership{Student Member,~IEEE,} 
		Ruisi He,~\IEEEmembership{Senior Member,~IEEE,}
		Minseok Kim,~\IEEEmembership{Senior Member,~IEEE,} 
		Inocent Calist, ~\IEEEmembership{Student Member,~IEEE,} 
        Mi Yang,~\IEEEmembership{Member,~IEEE,}
        and Ziyi Qi	

	\thanks{
%
An earlier version of this paper was presented in part at the 
IEEE International Conference on Communications (IEEE ICC-2026) \cite{zxj_icc}.

X. Zhang, R. He,  M. Yang and  Z. Qi are with the State Key Laboratory of Advanced Rail Autonomous Operation, the School of Electronics and Information Engineering, and the Frontiers Science Center for Smart High-speed Railway System,  Beijing Jiaotong University, Beijing 100044, China
(email: 23115029@bjtu.edu.cn; ruisi.he@bjtu.edu.cn; myang@bjtu.edu.cn; 22115006@bjtu.edu.cn).

M. Kim and I. Calist are with the Graduate School of Science and Technology, Niigata University, Niigata, Japan
(email: mskim@eng.niigata-u.ac.jp; f22l501j@mail.cc.niigata-u.ac.jp).

}
}
	\maketitle

\begin{abstract}
The deep integration of communication with intelligence and sensing, as a defining vision of 6G, renders environment-aware channel prediction a key enabling technology.
As a representative 6G application, vehicular communications require accurate and forward-looking channel prediction under stringent reliability, latency, and adaptability demands. 
Traditional empirical and deterministic models remain limited in balancing accuracy, generalization, and deployability, while the growing availability of onboard and roadside sensing devices offers a promising source of environmental priors.
This paper proposes an environment-aware channel prediction framework based on multimodal visual feature fusion. 
Using GPS data and vehicle-side panoramic RGB images, 
together with semantic segmentation and depth estimation, 
the framework extracts semantic, depth, and position features through a three-branch architecture and performs adaptive multimodal fusion via a squeeze-excitation attention gating module. 
For 360-dimensional angular power spectrum (APS) prediction, a dedicated regression head and a composite multi-constraint loss are further designed.
As a result, joint prediction of path loss (PL), delay spread (DS), azimuth spread of arrival (ASA), azimuth spread of departure (ASD), and APS is achieved.
Experiments on a synchronized urban V2I measurement dataset yield the best root mean square error (RMSE) of 3.26 dB for PL, RMSEs of 37.66 ns, 5.05$^\circ$, and 5.08$^\circ$ for DS, ASA, and ASD, respectively, and mean/median APS cosine similarities of 0.9342/0.9571, 
demonstrating strong accuracy, generalization, and practical potential for intelligent channel prediction in 6G vehicular communications.

\end{abstract}


\section{Introduction}

  \IEEEPARstart{T}{he} 
deep integration of communications and artificial intelligence (AI), together with the convergence of sensing and communication, has been identified by the ITU-R as a defining vision for 6G \cite{vspace0mmITUR2023, zhang2022}. 
Within this framework, accurate and intelligent channel prediction is regarded as a key enabler of 6G \cite{He2025}.
As a representative 6G scenario, vehicular communications are being driven by connected  and autonomous vehicles, imposing stringent requirements on reliability, latency, and adaptability.
However, high mobility, dynamic scatterers, and frequent non-line-of-sight (NLoS) transitions render wireless channels highly nonlinear and fast-varying, making accurate and robust prediction particularly challenging \cite{He2020}.

Conventional prediction methods are inherently constrained. 
Statistical approaches are strongly dependent on scenario-specific priors and thus suffer from limited generalization. 
Ray-tracing methods require accurate three-dimensional (3D) scene reconstruction and incur prohibitive complexity. 
Geometry-based stochastic models are efficient and easy to implement, but their idealized, scenario-specific assumptions limit their ability to capture dynamic vehicular channels \cite{he2026}.
Recently, with the widespread deployment of onboard and roadside sensors, such as  cameras and LiDAR, out-of-band environment sensing information has become increasingly available \cite{Huang2022}.
Wireless propagation is essentially determined by the physical environment, and many key environmental factors can be directly or indirectly observed by various sensing modalities.
Compared to conventional methods, such sensing data can provide informative priors for channel prediction without additional spectrum overhead, making environment-aware prediction a promising approach for improving accuracy, generalization, and deployment practicality \cite{bai2026}.

Among available sensing modalities, visual information is particularly attractive due to its low cost, practical accessibility, and strong ability to characterize propagation-relevant features in real time \cite{Tian2021}.
Since propagation-relevant factors, including the scale, location, and blockage relationships of scatterers, can be directly or indirectly inferred from visual data, and can now be efficiently extracted by mature computer vision (CV) techniques including  semantic segmentation and depth estimation,
vision-aided wireless communication has emerged as a promising research direction and has been increasingly validated as an effective means of enhancing wireless system performance \cite{Nishio2021}.

\subsection{Related Work} 
Existing studies on vision-aided wireless communication can be broadly categorized into two main directions, and the relevant work has been summarized in Table \ref{review}.
The first concerns vision-aided millimeter-wave (mmWave) beam management and blockage prediction, representing a mainstream application of visual information in wireless communications. 
For example, beam and blockage prediction are investigated in \cite{Charan2021,Yang2023} using simulated RGB images captured at the transmitter (Tx).
Studies such as \cite{imran2024, charan2025} further investigate beam prediction using measured multimodal dataset \cite{Alkhateeb2023} and RGB images captured at Tx.
Similarly, \cite{feng2025} introduces point clouds besides RGB images to achieve beam prediction in a simulated factory environment. 
Despite their effectiveness, these studies focus primarily on link-level optimization and provide limited insight into the fine-grained prediction of intrinsic channel characteristics required for system-level optimization.

\begin{table*}
	\belowrulesep=0pt
	\aboverulesep=0pt
	\renewcommand{\arraystretch}{1.2}   
	\setlength{\tabcolsep}{4pt}  
	\begin{center}
		\caption{Related Work on Environment-Aware Channel Prediction.}
		\label{review}
		\begin{threeparttable}
			\begin{tabular}{ c| c| c| c| c| c| c}
				\toprule
				\textbf{Reference}      & \textbf{Scenario} & \textbf{Freqency} & \textbf{Data Modality} & \textbf{Data Source }
				& \textbf{Sensor  Placement\tnote{*}} & \textbf{Target Metric}                      \\
				\midrule
				
				\hline
				\cite{Charan2021,Yang2023}     & V2I     & 28 GHz              & RGB Image & Simulation & Tx  & Beam \& blockage prediction \\
				\hline    
				\cite{imran2024, charan2025}    & V2I     & 60 GHz              & \makecell[c]{RGB Image} & Measurement & Tx  & Beam prediction          \\
				\hline    
				\cite{feng2025}       & Industry & 28 GHz         & RGB Image, point cloud & Simulation & Factory Ceiling  & Beam prediction    \\
				\hline 
				\cite{Nishio2019}     & Indoor  & 60 GHz              
				& Depth image 
				& Simulation & Tx \& Rx & Received Power   \\
				\hline
				\cite{Zhang2025b}     & V2I  & 60 GHz              
				& RGB image 
				& Measurement & Tx & Received Power   \\
				\hline    
				\cite{wang2025}       & V2I     & --                  & RGB image, point cloud, GPS data & Measurement & Tx  & PL                       \\
				\hline    
				\cite{wei2024}        & V2I     & 5.9 GHz \& 28 GHz   & RGB \& depth image,  point cloud & Simulation & Tx \& Rx  & PL\\
				\hline
				\cite{lu2025}         & V2I     & 28 GHz              & RGB \& depth image & Simulation & Tx \& Rx  & PL         \\
				\hline    
				\cite{Sun2024}        & A2G     & 28 GHz  & RGB \& depth image & Simulation & Tx (UAV-borne) & PL  \\
				\hline
				\cite{Zhang2025a}     & V2I     & 5.9 GHz             & RGB Image & Measurement & Tx  & PL, K-factor, RMS DS     \\
				\hline       
				\cite{zhou2025}       & Industry & 3.5 GHz \& 5.5 GHz  & RGB Image & Measurement & Tx \& Rx  & Received Power, RMS DS  \\  
				\hline
				\cite{xin2026}        & A2G     & 18 GHz   & \makecell[c]{ RGB image, 
					point cloud, GPS data} & Simulation & Tx (UAV-borne)  & CIR  \\
				\hline
				\cite{yin2026}        & Indoor  & 60 GHz             
				& RGB image, point cloud, GPS data 
				& Measurement & Rx  & RMS DS,  RMS AS   \\
				\bottomrule
			\end{tabular}
			\vspace{0.2em}
			\parbox{\linewidth}{\footnotesize
				\textit{Notes}:  
				* Tx and Rx denote the base-station and user terminals, respectively, consistent with the downlink configuration assumed in 3GPP channel models.
			}
		\end{threeparttable} 
	\end{center}
\end{table*}

The second direction concerns vision-aided propagation characteristic prediction.
Most existing work  focus on predicting single large-scale channel metrics, including received power and path loss (PL), from visual information. 
Typical examples include received-power prediction from time-series depth images acquired at both Tx and the receiver (Rx) in indoor environments \cite{Nishio2019},
received-power prediction in vehicle-to-infrastructure (V2I) settings using measured dataset \cite{Alkhateeb2023} and Tx-side RGB images \cite{Zhang2025b}, and PL prediction with additional point-cloud and GPS information \cite{wang2025,wei2024}.
Likewise, \cite{lu2025} and \cite{Sun2024} develop multimodal fusion models on the simulated dataset \cite{m3sc} for PL prediction in V2I and air-to-ground (A2G) scenarios, respectively.
Beyond scalar targets, only limited efforts address multidimensional channel characterization.
For example, \cite{Zhang2025a} jointly predicts PL, Ricean K-factor, and root mean square delay spread (RMS DS) from realistic Tx-side RGB images with instance segmentation, while \cite{zhou2025} estimates received power and RMS DS from real RGB images captured at both link ends in a factory environment.
In addition, \cite{xin2026} combines RGB images, point clouds, and GPS data to predict multipath amplitudes, delays, and angles, thereby enabling channel impulse response (CIR) reconstruction in an A2G scenario based on \cite{m3sc}. 
Similarly, \cite{yin2026} predicts RMS DS and RMS angle spread (AS) from measured multimodal data in indoor mmWave environments.

While these studies demonstrate the potential of visual or multimodal data for channel prediction, three key limitations remain. 
First, existing multimodal and multi-parameter frameworks are developed mainly for industrial or aerial scenarios, whose propagation characteristics differ markedly from those of V2I environments, limiting direct transferability. 
Moreover, most existing V2I studies rely on Tx-side visual inputs, although Rx-side sensing remains largely unexplored despite offering dynamic, receiver-aligned views that better reflect the actual propagation conditions than the fixed global perspective of Tx cameras. 
Second, most  methods are limited to scalar targets, such as PL and DS, and rarely address angular-domain characteristics, especially the 360$^\circ$ high-dimensional angular power spectrum (APS), whose structural and physical complexity makes it far more challenging than conventional scalar regression.
Third, most studies remains grounded in simulated environments and synthetic data, leaving systematic validation on real V2I measurement datasets still insufficient.

In summary, a unified multimodal visual feature fusion framework for the joint and accurate prediction of PL, DS, AS, and high-dimensional APS in real urban vehicular communications is still lacking, particularly one that can simultaneously ensure accuracy, generalization, and practical deployability.

\subsection{Contributions}
To address the above gaps, we propose a deep learning framework for channel prediction based on multimodal visual feature fusion for vehicle communication.
By jointly exploiting onboard panoramic RGB images and GPS data, together with semantic segmentation and  depth estimation, the proposed framework enables the joint prediction of PL, DS, azimuth spread of arrival (ASA), azimuth spread of departure (ASD), and APS.
While our earlier work \cite{zxj_icc} reports preliminary results limited to PL prediction, the present study substantially extends it through a more robust architecture and more comprehensive experimental validation.
The main contributions of this paper are summarized as follows:
\begin{itemize}
	\item[$\bullet$]  
	A multimodal environment-aware channel prediction framework is established by defining semantic, depth, and position features as inputs, formulating the joint prediction of five representative channel parameters under physical constraints, and designing a weighted joint loss for heterogeneous targets.
	\item[$\bullet$]  
	An adaptive multimodal visual feature fusion network is developed by encoding semantic, depth, and positional inputs into a unified latent space, enabling attention-guided cross-modal fusion, and further introducing a dedicated regression prediction head with a composite multi-constraint loss for APS prediction.
	\item[$\bullet$]  
	A real-world multimodal dataset is established from systematic urban V2I measurements using synchronized channel sounding, Rx-side panoramic imaging, and GPS, with one-to-one cross-modal correspondence ensured through calibration, preprocessing, semantic/depth generation, spatiotemporal alignment, and sample filtering.
	\item[$\bullet$]  
    Extensive experiments, including modal ablation, dynamic-scatterer analysis, backbone comparison, and APS-oriented evaluation, validate the proposed framework and demonstrate its advantages in accuracy, generalization, and practical applicability.
\end{itemize}

The remainder of this paper is organized as follows. 
Section II presents system model and problem formulation. 
Section III   introduces the proposed  multimodal  visual feature fusion framework.
Section IV describes 
the construction of multimodal dataset. 
Section V reports the experimental results and corresponding analysis.
Section VI draws the conclusions.

\section{System Model and Methodology}
\subsection{System Model}
We consider a  V2I communication system operating in the urban environment.
The system consists of a static roadside base station (BS) acting as Tx and a mobile vehicle acting as Rx equipped with a panoramic camera and a GPS receiver.
Panoramic camera captures a 360\degree\ field of view (FoV) of the surrounding propagation environment in real time during communication. 
GPS receiver collects the geographic coordinates of both Tx and Rx. 
To capture the electromagnetic scattering features of environment, we leverage multimodal environmental features, including visual and location modalities. 
Formally, we define them at  $t$ as follows:

\textbf{1) Visual Modality:} 
The vehicle captures RGB panoramic images, denoted as $\mathbf{I}_{\text{RGB}}[t] \in \mathbb{R}^{C \times H \times W  }$, where $C$, $H$,  and $W$ denote the number of color channels (typically $C=3$ for RGB images) of the image, height,  and width, respectively. 
To extract features relevant to radio propagation, we utilize two CV techniques to generate two derived modalities:
\begin{enumerate}
	\item[a)] \textit{Semantic View:} A semantic segmentation image $\mathbf{S}[t] \in \mathbb{R}^{C_{\text{semantic}} \times H \times W  }$, where each pixel is classified into categories 
	that exhibit distinct dielectric properties.
	\item[b)] \textit{Depth View:} A depth image $\mathbf{D}[t] \in \mathbb{R}^{C_{\text{depth}} \times H \times W}$, representing the relative distance from Rx to surrounding scatterers, which is crucial for estimating path delays.
\end{enumerate}

\textbf{2) Location Modality:} The geometric relationship is captured by   coordinate vectors $\mathbf{G}_{\text{T}}(t), \mathbf{G}_{\text{R}}(t) \in \mathbb{R}^2$, which denote the coordinates of  Tx and Rx at time instance $t$, respectively.

Thus, the holistic environmental state space at time $t$ is defined as $\Omega [t] = \{ \mathbf{S}[t], \mathbf{D}[t], \mathbf{G}_{\text{T}}[t],  \mathbf{G}_{\text{R}}(t)\}$.

\subsection{Prediction Targets}
We focus on the joint prediction of multiple  channel  parameters, including PL, DS, ASA, ASD, and APS. 
The five target parameters to be predicted are collectively denoted as the output set $\mathbf{Y}[t]$. Each parameter is formally defined with its physical meaning and mathematical expression as follows:

\textit{1) PL:} $\widehat{L}[t] \in \mathbb{R}^+$, a non-negative scalar representing the attenuation of  communication signals during propagation from Tx to Rx, with the unit of decibel (dB). It is determined by the relative distance, environmental occlusion, and other factors.

\textit{2) DS:} $\widehat{\tau}[t] \in \mathbb{R}^+$, a non-negative scalar, with the unit of nanosecond (ns). 
It reflects the time-domain dispersion degree of MPCs.

\textit{3) ASA:} $\widehat{\alpha}[t] \in \mathbb{R}^+$, a non-negative scalar representing the  azimuth spread of signals at Rx, with the unit of degree (\degree). 
It describes the spread range of  arrival directions of  MPCs.

\textit{4) ASD:} $\widehat{\beta}[t] \in \mathbb{R}^+$, a non-negative scalar representing the  azimuth spread of signals at Tx, with the unit of degree (\degree). 
It describes the spread range of  departure directions of MPCs.

\textit{5) APS:} $\widehat{\mathbf{P}}[t] \in \mathbb{R}^{360 \times 1}$, a 360-dimensional non-negative vector, where each element $\widehat{P}_k[t] \in [0,1]$ ($k=0,1,\dots,359$) represents the normalized power of  received signals at the angle $k^\circ$ (corresponding to the 360\degree\ FoV of panoramic camera). 
APS is normalized to the range $[0,1]$ with the peak power as the reference, i.e., $\max(\widehat{\mathbf{P}}[t]) = 1$.

Thus, the output set $\mathbf{Y}[t]$ is formally expressed as:
\begin{equation}
	\label{math3}
	\mathbf{Y}[t] = \left\{ \widehat{L}[t], \widehat{\tau}[t], \widehat{\alpha}[t], \widehat{\beta}[t], \widehat{\mathbf{P}}[t] \right\}.
\end{equation}

\subsection{Problem Formulation}
The problem addressed in this work is to design a deep learning model with parameter set $\Theta$, which learns a mapping function $f_\Theta$ from  $\Omega[t]$ to  $\mathbf{Y}[t]$, such that the predicted values of channel parameters are as close as possible to their true values, while satisfying the corresponding physical constraints.


\subsubsection{Dataset Definition}
Let $D = \left\{ (\Omega_m, \mathbf{Y}_m^\star) \right\}_{m=1}^M$ denote the labeled dataset collected from actual V2I propagation environment, where $M$ is the total number of snapshots in the dataset. For each snapshot $m$ ($m=1,2,\dots,M$):
\begin{itemize}
	\item $\Omega_m = \left\{ \mathbf{S}_m, \mathbf{D}_m, \mathbf{G}_{T,m}, \mathbf{G}_{R,m}\right\}$ is the multimodal input feature set of the $m$-th sample, where $\mathbf{S}_m$, $\mathbf{D}_m$, $\mathbf{G}_{T,m}$, and $\mathbf{G}_{R,m}$ are  semantic image, depth image, Tx GPS coordinate, and Rx GPS coordinate, respectively.
	\item $\mathbf{Y}_m^\star = \left\{ L_m^\star, \tau_m^\star, \alpha_m^\star, \beta_m^\star, \mathbf{P}_m^\star \right\}$ is the true value set of the five target parameters for the $m$-th sample, where $L_m^\star$, $\tau_m^\star$, $\alpha_m^\star$, $\beta_m^\star$, and $\mathbf{P}_m^\star$ are the true PL, DS, ASA, ASD, and APS of the $m$-th sample, respectively. 
\end{itemize}

\subsubsection{Loss Function Design}
To train the multi-target prediction model, target-specific loss functions are adopted for different channel parameters. 
For a generic scalar target $x\in{L,\tau,\alpha,\beta}$, the mean squared error (MSE) is adopted as
\begin{equation}
	Loss(\Theta)={\frac{1}{M}\sum_{m=1}^{M}\left(x_m^\star-\widehat{x}_m\right)^2},\quad x\in{L,\tau,\alpha,\beta},
\end{equation}
where $ x_m^\star $ and $\widehat{x}_m$ denote the ground-truth and predicted values of the target parameter for the $m$-th sample, respectively.
For APS prediction, a dedicated composite loss function is designed to address its unique challenges, including angular periodicity, shape fitting requirements, and physical constraints. The detailed formulation and design rationale of this loss function are elaborated in Section III-C.
\subsubsection{Optimization Objective with Physical Constraints}
The ultimate goal   is to learn the optimal model parameter set $\Theta^\star$ by minimizing the  loss function $Loss(\Theta)$, while ensuring that the predicted values of all target parameters satisfy their corresponding physical constraints. The optimization problem can be formally formulated as:
\begin{equation}
	\label{math6}
	\Theta^\star = \mathop{\arg\min}_{\Theta} Loss(\Theta),
\end{equation}
subject to the following physical constraints for all samples:
\begin{enumerate}
	\item Non-negativity constraints for scalar parameters: $\widehat{L}_m \geq 0$, $\widehat{\tau}_m \geq 0$, $\widehat{\alpha}_m \geq 0$, $\widehat{\beta}_m \geq 0$;
	\item APS constraints: $\widehat{P}_{m,k} \in [0,1]$ for all $k=0,1,\dots,359$, and $\max(\widehat{\mathbf{P}}_m) = 1$.
\end{enumerate}
These constraints are embedded into the model design and training process to ensure that the predicted results are physically meaningful and consistent with  channel characteristics.

\begin{figure*}[!t]
	\centering
	\includegraphics[width=.90\textwidth]{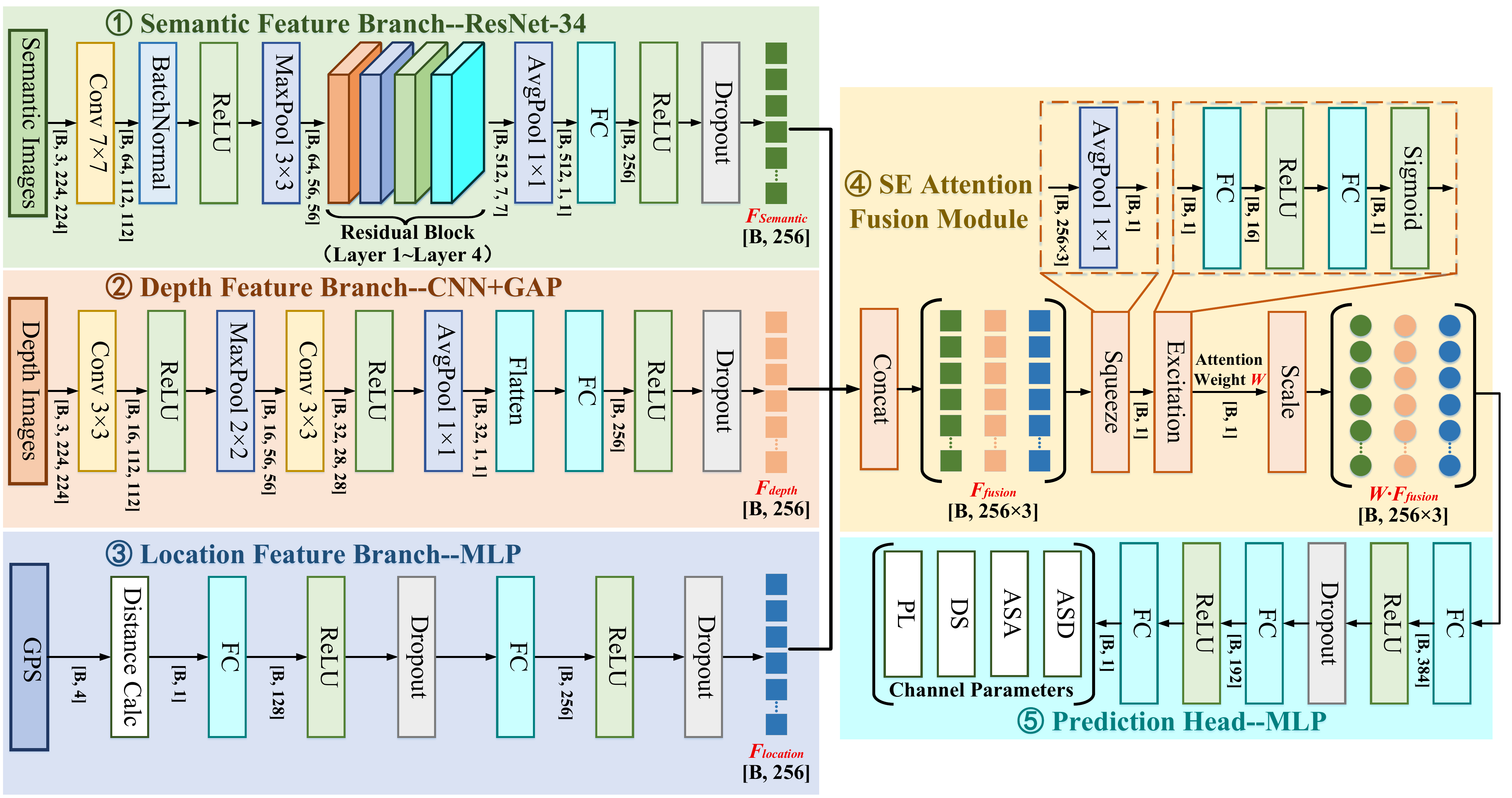}%
	\caption{The proposed channel prediction model architecture based on visual sensing and multimodal feature fusion.}                    
	\label{framework}
\end{figure*}

\section{Multimodal Visual Feature Fusion Based Channel  Prediction Model}
\subsection{Model Architecture}
In this study, we propose a channel prediction framework for V2I communications based on multimodal visual feature fusion. By jointly leveraging semantic, depth-related, and geometric location information, the framework performs channel prediction through three key stages: multi-branch feature extraction, squeeze-and-excitation (SE)-attention-based gated fusion, and multilayer perceptron (MLP)-based regression, as shown in Fig. \ref{framework}.
Specifically, heterogeneous inputs are encoded into a unified latent space, adaptively fused, and then mapped to the target channel parameters.

\textbf{1) Feature Extraction:}
The model employs a three-branch multimodal encoder to separately extract propagation-relevant features from each modality.

\textit{Semantic Feature Branch:} A pretrained ResNet-34 backbone is adopted to encode semantic segmentation images resized to $[B,3,224,224]$, where $B$ denotes the batch size. 
To improve training stability and reduce overfitting under limited data, the initial \textit{Conv}, \textit{BatchNorm}, and residual $Layer~1$-$Layer~2$ are frozen, while only $Layer~3$, $Layer~4$, and a newly introduced feature mapping head are fine-tuned.
Detailed network structures of residual block can be found in \cite{He2016}.
Specifically, the original classifier is replaced with an ``fully connected (FC)$\rightarrow$ReLU$\rightarrow$Dropout'' module, which projects the 512-dimensional backbone output into a unified 256-dimensional semantic feature vector ${F_{{\rm semantic}}}\in\mathbb{R}^{B\times256}$.

\textit{Depth Feature Branch:} Depth images resized to $[B,3,224,224]$ are encoded by a lightweight convolutional neural network (CNN). Local spatial features are first extracted, then compressed by global average pooling (GAP), and finally projected through a FC layer into a unified 256-dimensional depth feature vector ${F_{{\rm depth}}}\in\mathbb{R}^{B\times256}$. This branch provides a low-cost global representation of the depth modality, complementing environment-related 3D structural cues for propagation characterization.

\textit{Location Feature Branch:} To encode communication geometry, the longitude and latitude of Tx and Rx are first converted into a 1-dimensional (1D) geodetic distance, which is then fed into a two-layer MLP to generate a 256-dimensional location feature vector ${F_{{\rm location}}}\in\mathbb{R}^{B\times256}$. 
In this way, location modality is projected into the same feature space as the semantic and depth branches for subsequent fusion.

\textbf{2) Adaptive Feature Fusion Based on SE Attention:}
A gated SE attention module is introduced for adaptive multimodal fusion. Specifically, branch features are first concatenated along the channel dimension to form the fused representation ${F_{{\rm fusion}}}$, which is then globally pooled and passed through a lightweight gating network to generate attention weights ${W}$. The fused features are subsequently rescaled by broadcast multiplication. Since ${W}\in\mathbb{R}^{B\times1}$, the proposed strategy performs sample-level gating, i.e., applying a shared scaling factor to each sample, thereby suppressing noisy or low-confidence samples and improving training stability and cross-scenario robustness.

\textbf{3) Prediction Regression Head:}
An MLP is used to regress the fused features into scalar channel parameters, including PL, DS, ASA, and ASD, by progressively reducing the feature dimension and learning the nonlinear mapping from high-dimensional representations to propagation characteristics. Owing to the 360-dimensional normalized structure of APS, the regression head in Fig. \ref{framework} is not directly applicable, and a dedicated prediction architecture is presented separately in Section III-B.



\begin{figure}[!t]
	\centering
	\includegraphics[width=.40\textwidth]{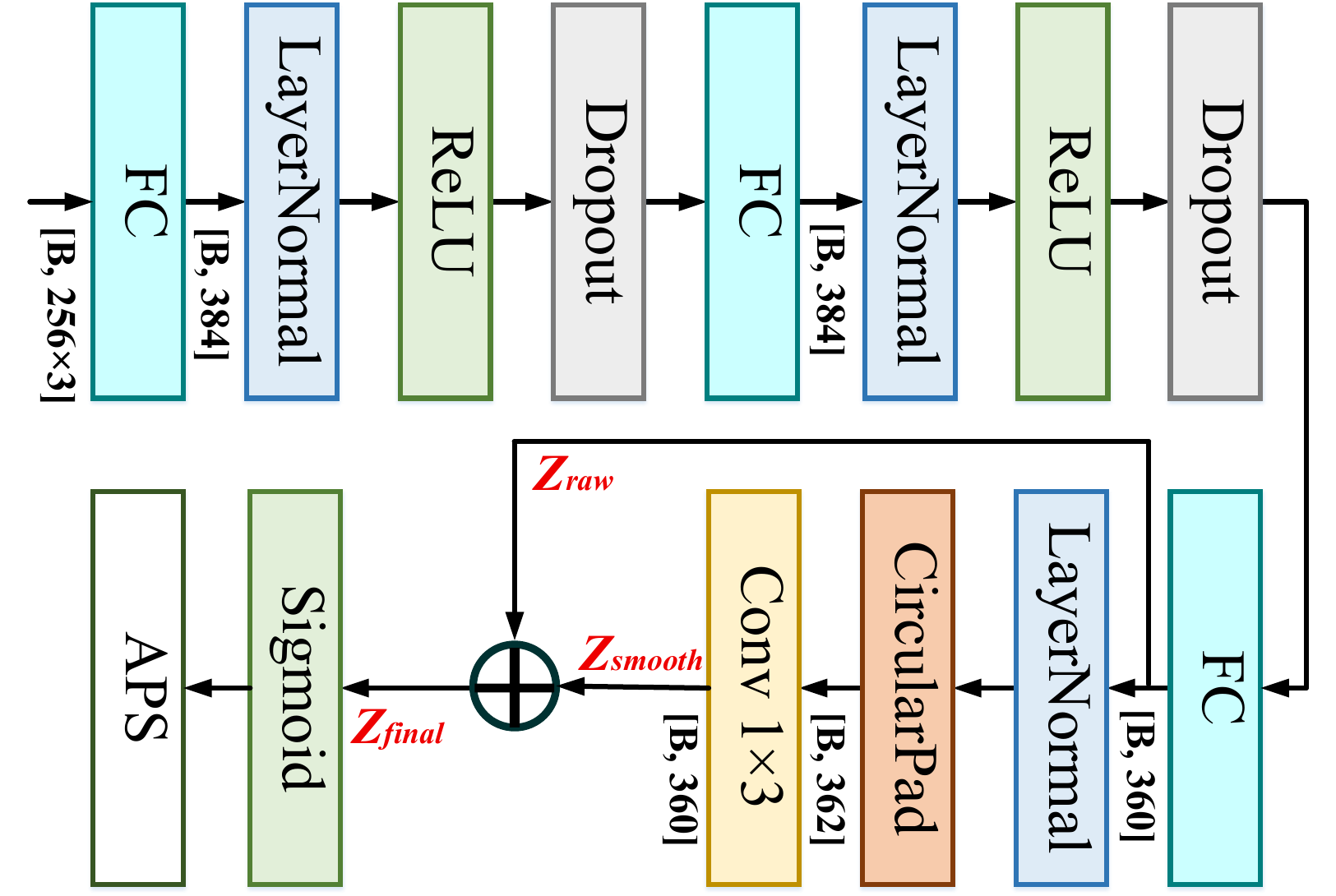}%
	\caption{Regression head architecture specifically designed for APS prediction.}                    
	\label{aps_head}
\end{figure}

\subsection{Prediction Head Design For APS Prediction }
This module is redesigned from a single-output linear layer into a dedicated regression head for 360-dimensional (360D) APS prediction. As shown in Fig. \ref{aps_head}, the fused features are first projected by an MLP with \textit{LayerNorm} and a reduced dropout ratio into a 360D intermediate representation, where \textit{LayerNorm} improves training stability and the lower dropout preserves fine-grained low-power details. To exploit the inherent 360$^\circ$ periodicity of APS, a 1D recurrent convolution module with \textit{CircularPad} is further introduced, preserving the continuity between 0$^\circ$ and 359$^\circ$ while applying mild local smoothing. A residual fusion mechanism then combines  raw features $Z_{\text{raw}}$ and  smoothed features $Z_{\text{smooth}}$ in a ratio of 0.2:0.8 to balance local detail preservation and global smoothness. Finally, a sigmoid activation constrains the output to $[0,1]$, ensuring consistency with the normalized APS definition.



\subsection{Loss Function Design For APS Prediction }

The loss function largely determines the fitting accuracy of  APS.
Unlike scalar regression, APS learning must address three challenges:
(i) preserving the global curve geometry (e.g., main-lobe location and sidelobe structure),
(ii) preventing the low-power region from being dominated by the high-power main lobe,
and (iii) enforcing physical consistency, including amplitude bounds in $[0,1]$ and and total-power conservation. 
To this end,  a composite loss with multi-dimensional constraints is designed as 
\begin{equation}
	\label{math9}
	\zeta_{\mathbf{P}}(\mathbf{P}_m^\star, \widehat{\mathbf{P}}_m) = \mathcal{L}_{\text{shape}} + \omega_{\text{mse}} \cdot \mathcal{L}_{\text{WMSE}} + \omega_{\text{L1}} \cdot \mathcal{L}_{\text{WL1}} + \omega_{\text{tp}} \cdot \mathcal{L}_{\text{RTP}},
\end{equation}
where $\mathcal{L}_{\text{shape}}$ denotes the shape similarity constraint loss, $\mathcal{L}_{\text{WMSE}}$ is the Weighted Mean Squared Error (WMSE) for point-wise amplitude constraint, $\mathcal{L}_{\text{WL1}}$ is the Weighted L1 Loss for amplitude overshoot suppression, and $\mathcal{L}_{\text{RTP}}$ is the Relative Total Power Loss for global amplitude regularization. 

\paragraph{Shape Similarity Constraint Loss ($\mathcal{L}_{\text{shape}}$)}
To avoid the problem that point-wise loss ignores the global shape, the cosine similarity loss is adopted as the core shape constraint:
\begin{equation}
	\label{math10}
	\mathcal{L}_{\text{shape}} = 1 - \frac{1}{M} \sum_{m=1}^M \cos\theta(\mathbf{P}_m^\star, \widehat{\mathbf{P}}_m),
\end{equation}
where $\cos\theta(\mathbf{P}_m^\star, \widehat{\mathbf{P}}_m)$ is the cosine similarity between $\mathbf{P}_m^\star$ and $\widehat{\mathbf{P}}_m$, formally defined as
\begin{equation}
	\label{math11}
	\cos\theta(\mathbf{P}_m^\star, \widehat{\mathbf{P}}_m) = \frac{\mathbf{P}_m^\star \cdot \widehat{\mathbf{P}}_m}{\|\mathbf{P}_m^\star\|_2 \cdot \|\widehat{\mathbf{P}}_m\|_2 + \epsilon}
\end{equation}
with $\epsilon=10^{-8}$ to avoid division by zero. 
Since cosine similarity depends only on the included angle between vectors, it emphasizes global shape consistency rather than absolute amplitude. 
Accordingly, the weight of $\mathcal{L}_{\text{shape}}$ is fixed to 1, reflecting the priority of shape fitting in APS prediction.

\paragraph{Low-Power Region Weighting Strategy}
A critical challenge in APS prediction is that the high-power main lobe (amplitude $\geq$ 0.5) can dominate the loss, causing insufficient fitting of low-power regions (amplitude $\textless$ 0.5), such as minor side lobes and local fluctuations. 
To alleviate this imbalance, a power-threshold-based weighting strategy is introduced:
\begin{equation}
	\label{math12}
	w_{m,k} = 1 + (\omega_{\text{low}} - 1) \cdot \mathbb{I}(P_{m,k}^\star < \tau),
\end{equation}
where $w_{m,k}$ is the  weight for the $k$-th angular point of the $m$-th sample,
$\mathbb{I}(\cdot)$ is the indicator function, 
$\tau=0.5$ is the power threshold,
and $\omega_{\text{low}}=8$ is the amplification factor for the low-power region.
This design increases the contribution of low-power samples and encourages the model to capture their fine-grained structures. 
The value of  $\tau=0.5$ and  $\omega_{\text{low}}=8$ are determined empirically to balance main-lobe fitting and low-power detail preservation.

\paragraph{Weighted Dual Amplitude Constraint ($\mathcal{L}_{\text{WMSE}} + \mathcal{L}_{\text{WL1}}$)}
Based on the low-power weighting strategy, a dual amplitude constraint combining is adopted.
$\mathcal{L}_{\text{WMSE}}$ provides precise point-wise amplitude fitting, particularly for the main lobe:
\begin{equation}
	\label{math13}
	\mathcal{L}_{\text{WMSE}} = \frac{1}{M \times 360} \sum_{m=1}^M \sum_{k=0}^{359} w_{m,k} \cdot (P_{m,k}^\star - \widehat{P}_{m,k})^2.
\end{equation}
$\mathcal{L}_{\text{WL1}}$ improves robustness to overshoot and outliers, thereby stabilizing the fitting of both main lobe and low-power regions:
\begin{equation}
	\label{math14}
	\mathcal{L}_{\text{WL1}} = \frac{1}{M \times 360} \sum_{m=1}^M \sum_{k=0}^{359} w_{m,k} \cdot |P_{m,k}^\star - \widehat{P}_{m,k}|.
\end{equation}
Both losses share the same low-power weighting mechanism. 
Their weights are set to $\omega_{\text{mse}}=0.065$ and $\omega_{\text{L1}}=0.025$, ensuring that amplitude constraint is auxiliary to shape constraint.

\paragraph{Relative Total Power Constraint Loss ($\mathcal{L}_{\text{RTP}}$)}
To ensure that the global total power of $\widehat{\mathbf{P}}_m$ is consistent with $\mathbf{P}_m^\star$, $\mathcal{L}_{\text{RTP}}$ is designed to avoid loss imbalance caused by differences in total power across different samples:
\begin{equation}
	\label{math15}
	\mathcal{L}_{\text{RTP}} = \frac{1}{M} \sum_{m=1}^M \frac{\left| \sum_{k=0}^{359} P_{m,k}^\star - \sum_{k=0}^{359} \widehat{P}_{m,k} \right|}{\sum_{k=0}^{359} P_{m,k}^\star + \epsilon}.
\end{equation}
Compared with absolute total power loss, 
$\mathcal{L}_{\text{RTP}}$ measures relative deviation and is therefore less sensitive to sample-wise power variation.
The weight $\omega_{\text{tp}}=0.01$ is chosen to provide a mild global regularization without interfering with shape and point-wise fitting.

%

\section{Experimental Setup}
High-quality datasets are critical to vision-aided channel prediction, as real measured visual data preserve authentic environmental characteristics and signal dynamics that simulated data often fail to capture. Therefore, extensive real-world V2I measurements are conducted to build a dedicated dataset for training and validation.
In this section, we detail the procedures for measurement campaign, dataset generation  and post-processing, and model training.

\begin{figure}[!t]
	\centering
	\includegraphics[width=.42\textwidth]{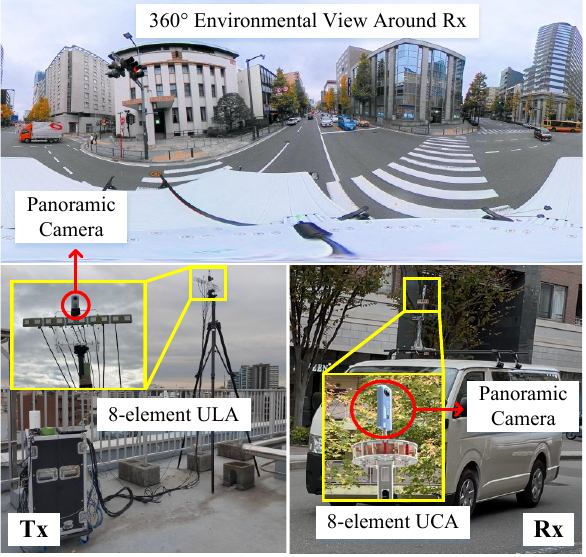}%
	\caption{Measurement system.
	}                    
	\label{system}
\end{figure}

\subsection{Measurement Campaign}
A custom-designed multiple-input multiple-output (MIMO) channel sounder is utilized to perform the channel measurements \cite{kim2012}. 
The system integrates high-stability RF frequency conversion modules with a dedicated antenna array optimized for  4.85 GHz.
An unmodulated Newman phase multi-tone signal is adopted as the sounding waveform, enabling an instantaneous bandwidth of 99.9 MHz, a fine delay resolution of 10 ns, a tone spacing of 195 kHz, and a measurable delay span of up to 5.12 \textmu s.
Both the Tx and Rx are equipped with 8-element uniform linear array (ULA) and uniform circular array (UCA), respectively, enabling an 8 × 8 MIMO configuration for wideband double-directional channel characterization. 
The half-power beamwidth (HPBW) of each element in the horizontal plane is 90\degree\ for ULA and 75\degree\ for UCA, with corresponding antenna gains of approximately 4 dBi and 6.5 dBi. 
Two panoramic cameras are mounted directly above Tx and Rx antennas to synchronously capture 360\degree\  environmental images without interfering with channel sounding process, as illustrated in Fig. \ref{system}. 
The detailed system parameters are summarized in Table \ref{Parameters}.

\begin{figure*}[!t]
	\centering
	
	\subfloat[]{\includegraphics[width=.24 \textwidth]{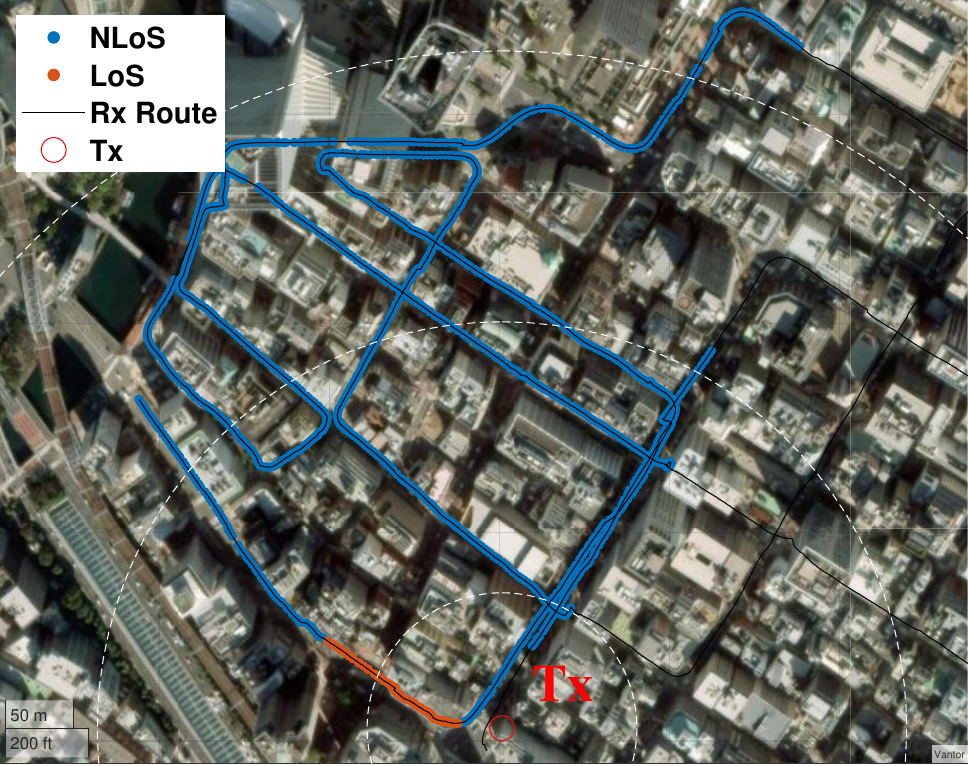}  \hspace{0mm}
		\label{R3_map2}}
	\subfloat[]{\includegraphics[width=.24 \textwidth]{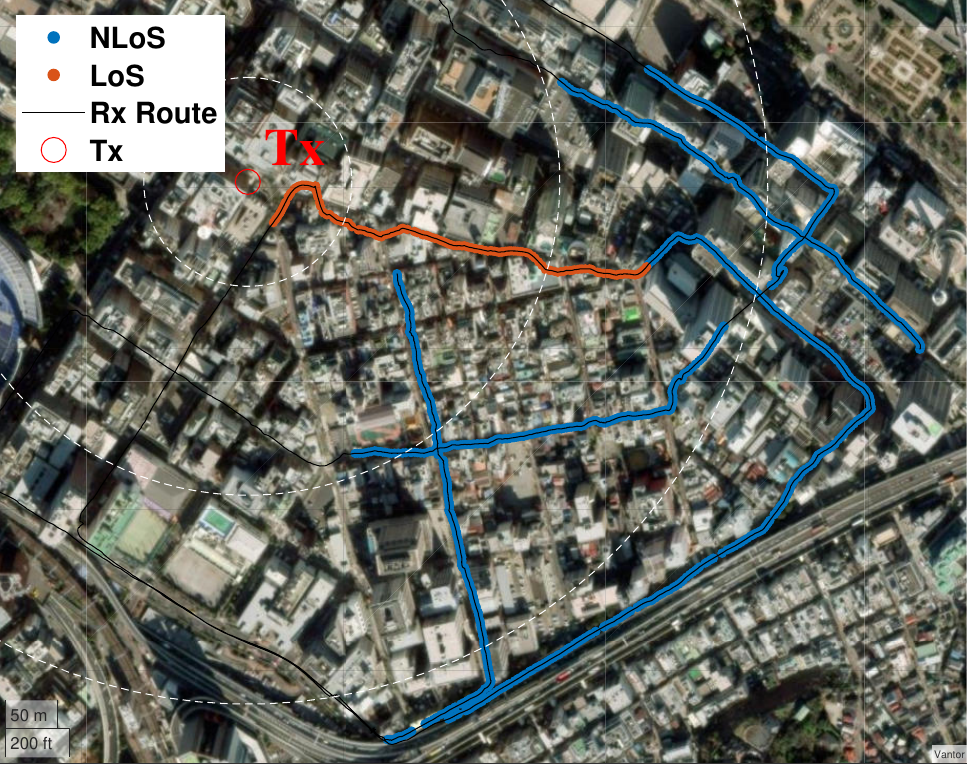}  \hspace{0mm}
		\label{R4_map2}}
	\subfloat[]{\includegraphics[width=.24 \textwidth]{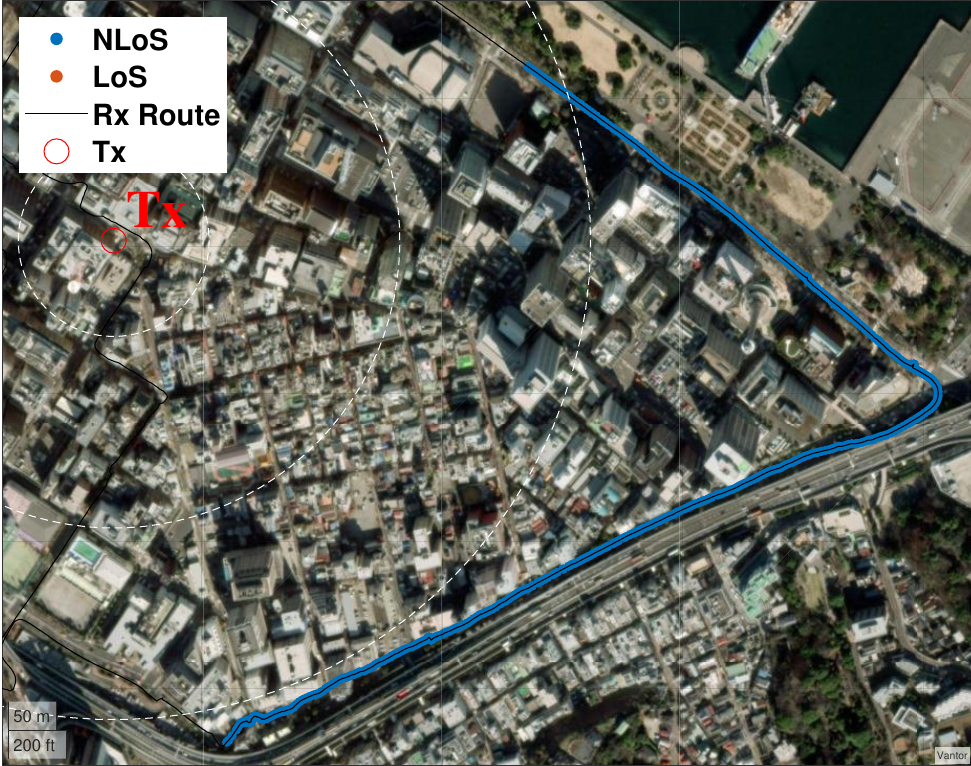}  \hspace{0mm}
		\label{R5_map2}}
	\subfloat[]{\includegraphics[width=.24 \textwidth]{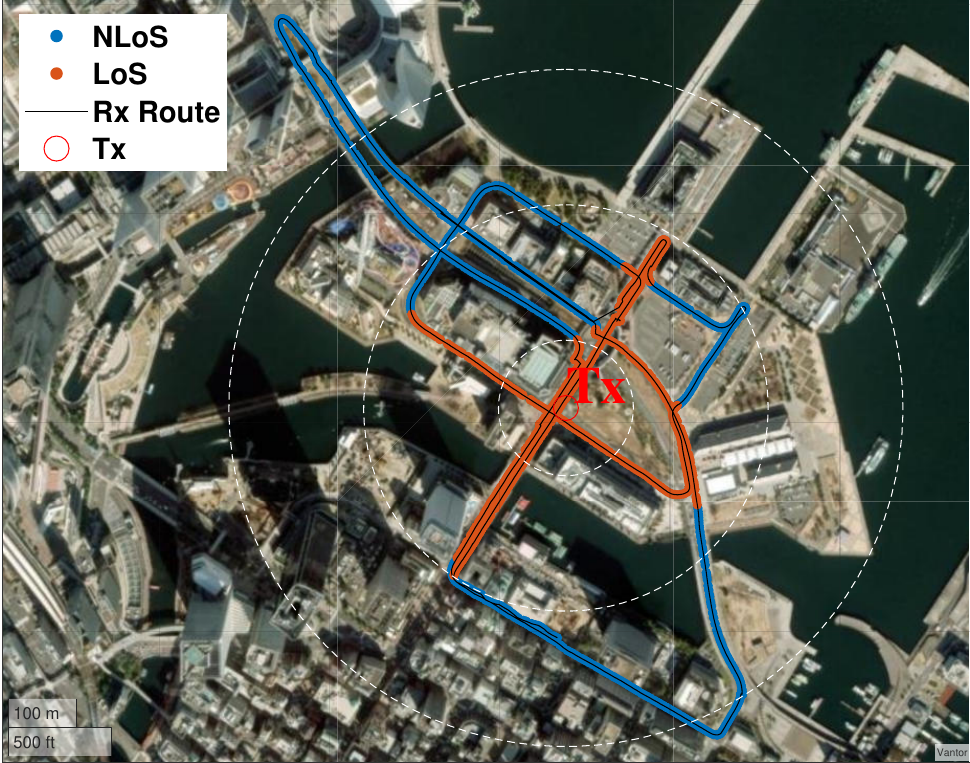}  
		\label{R6_map2}}
	
	\caption{Measurement trajectories along three different Areas in  urban scenarios.
		(a) \textit{Area 1};
		(b) \textit{Area 2};
		(c) \textit{Area 3};
		(d) \textit{Area 4}.
	}
	\label{scenarios}
\end{figure*}

To obtain sufficient training data, extensive V2I measurements are conducted across four urban areas in Yokohama, Japan. 
The BS is deployed at heights of 33 m, 34 m, 34 m, and 3 m in \textit{Areas 1-4}, respectively, with \textit{Areas 1-3} corresponding to urban macro (UMa) scenarios and \textit{Area 4} to an urban micro (UMi) scenario. 
In all cases, the ULA maintains a fixed azimuth coverage of $-50^\circ$ to $+50^\circ$. 
The Rx, mounted on a moving vehicle, carries the UCA and a panoramic camera on the rooftop at 2.7 m, while precise GPS information is available at both link ends. 
The vehicle traverses the measurement routes at speeds below 20 km/h, covering both LoS and NLoS segments, as shown in Fig. \ref{scenarios}. 
Channel snapshots at 4.85 GHz are recorded every 0.5 s, yielding 3,000 snapshots per area. Collectively, these routes capture representative urban street-canyon environments with dense buildings, vehicular and pedestrian traffic, and seasonal vegetation, thereby providing a rich joint channel–image dataset in real urban V2I scenarios.

\begin{table}[!t]
	\belowrulesep=0pt
	\aboverulesep=0pt
	\renewcommand{\arraystretch}{1.2}
	\centering
	\caption{Key Parameters of  Measurement Campaign}
	\label{Parameters}
	\begin{tabular}{c|c}
		\toprule
		\textbf{Parameters} & \textbf{Description}  \\
		\midrule
		Carrier Frequency & 4.85 GHz\\
		Bandwidth & 100 MHz \\
		No. Subcarriers & 510 \\
		Subcarrier Spacing & 195 KHz \\
		Sampling Rate & 800 MSa/s \\
		Delay Resolution & 10 ns \\
		Delay Span & 5.12 \textmu s \\
		Tx Power & 26 dBm/antenna \\
		Tx/Rx Array & 8-element ULA / UCA \\
		HPBW & \makecell[c]{ULA: 75\degree  (H) / 70\degree (V), UCA: 90\degree (H/V)} \\
		Antenna Gain & UCA: 6.5 dBi, ULA: 4 dBi \\
		Tx Height & 33 m (Area 1), 34 m (Areas 2\&3), 3 m (Area 4)  \\
		Rx Height  & 2.7 m  \\
		Vehicle Speed & $\leq$ 20 km/h \\
		
		\bottomrule
	\end{tabular}
\end{table}

\subsection{Dataset Generation and  Post-processing}
\begin{figure}[!t]
	\centering
	\includegraphics[width=.40\textwidth]{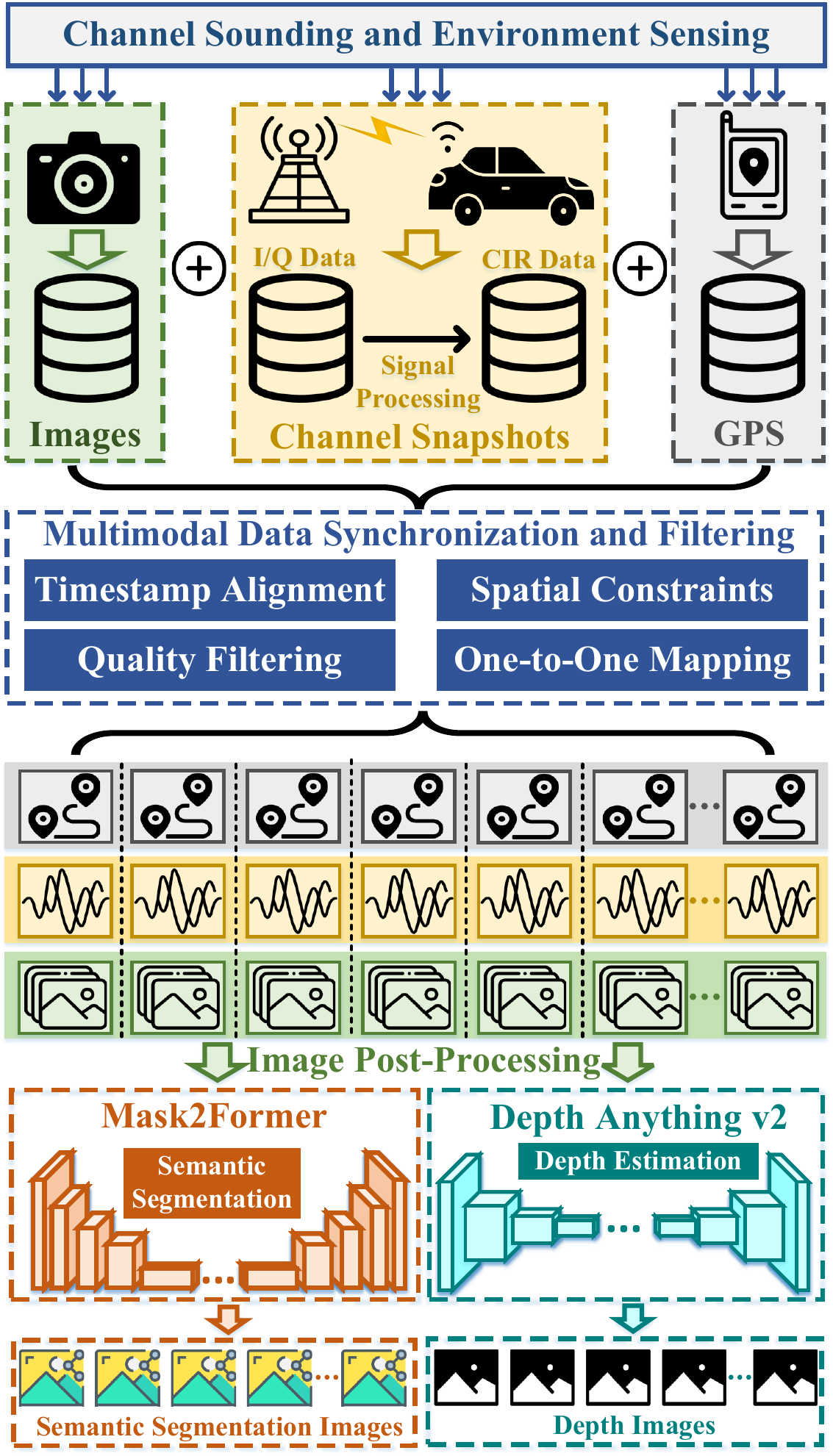}%
	\caption{Block diagram of the dataset generation pipeline.
	}                    
	\label{dataset}
\end{figure}

Three data modalities are acquired during measurements: channel sounding data, panoramic image data, and GPS data. 
However, these raw data cannot be directly used in the proposed model and must undergo a sequence of post-processing procedures to form a dataset suitable for training, as shown in Fig. \ref{dataset}. 
The processing pipeline  are described in detail below.

\subsubsection{Channel Data  Post-processing}
The first step is system calibration, including back-to-back (B2B) measurement and antenna-pattern calibration. 
In the B2B test, Tx and Rx are directly connected through RF cables and attenuators to compensate for the frequency-dependent amplitude response of  RF front-end, while Tx/Rx array radiation patterns are calibrated to reduce pattern-induced distortions for accurate channel-parameter extraction. Details of the calibration procedure are provided in \cite{calist2025}.

The measured MIMO channel impulse response is the convolution of underlying propagation channel and hardware response, and it is commonly modeled as the linear superposition of 
$L$ plane-wave MPCs. 
By incorporating the Tx/Rx array radiation patterns $\boldsymbol{a}_{T}\left({\varphi _{{\rm{T}},l}}\right)$ and $\boldsymbol{a}_{R}\left({\varphi _{{\rm{R}},l}}\right) $, and the autocorrelation function of  sounding waveform $a_{u}$, the measured MIMO channel response is expressed as:
\begin{equation}
h_{\text{MIMO}}(\tau)=\sum_{l=1}^{L} \gamma_{l} a_{u}\left(\tau-\tau_{l}\right) \boldsymbol{a}_{\rm{R}}\left({\varphi _{{\rm{R}},l}}\right) \boldsymbol{a}_{\rm{T}}^{T}\left({\varphi _{{\rm{T}},l}}\right),
\end{equation}
where $\gamma_{l}$, $\tau_{l}$, ${\varphi _{{\rm{T}},l}}$,  ${\varphi _{{\rm{R}},l}}$
represent the complex path weight, delay, angle of departure (AoD), and angle of arrival (AoA) of the $l$-th path, respectively.
To mitigate distortions introduced by system hardware and  extract MPC parameters, the Spatial Alternating Generalized Expectation-Maximization (SAGE) algorithm is employed. 
Accordingly, the estimated parameter set is obtained as $\{\hat \gamma_{l}$, $\hat \tau_{l}$, $\hat {\varphi _{{\rm{T}},l}}$,  $\hat {\varphi _{{\rm{R}},l}}\}$.

The large-scale parameters, including PL, DS, ASD, and ASA, are calculated from the above extracted MPC parameters. 
To suppress small-scale fading, the total received power is computed via incoherent summation of  extracted  MPC powers. 
PL of each snapshot is then obtained from the aggregate path gain of all extracted MPCs as:
\begin{equation}
{\rm{PL}} =  - 10{\log _{10}}\sum\limits_{l = 1}^L {{{\left| {{{\hat \gamma }_l}} \right|}^2}}.
\end{equation}
The temporal dispersion is characterized by  RMS DS, computed as the square root of the second central moment:
\begin{equation}
{\rm{DS}} = \sqrt {\sum\limits_{l = 1}^L {\hat \tau _l^2} {{\left| {{{\hat \gamma }_l}} \right|}^2}/\sum\limits_{l = 1}^L {{{\left| {{{\hat \gamma }_l}} \right|}^2}}  - {{\left( {\sum\limits_{l = 1}^L {\hat \tau _l^2} {{\left| {{{\hat \gamma }_l}} \right|}^2}/\sum\limits_{l = 1}^L {{{\left| {{{\hat \gamma }_l}} \right|}^2}} } \right)}^2}}. 
\end{equation}
Similarly, the azimuth angular spread at Tx and Rx is computed from the extracted MPCs as
\begin{equation}
{\rm{ASD/ASA}} = \sqrt { - 2\ln \left| {\sum\limits_{l = 1}^L {\exp (j{{\hat \varphi }_{\rm{T}(\rm{R}),l}})} {{\left| {{{\hat \gamma }_l}} \right|}^2}/\sum\limits_{l = 1}^L {{{\left| {{{\hat \gamma }_l}} \right|}^2}} } \right|}.
\end{equation}
Accordingly, taking angle of arrival as an example, its APS can be calculated using the following formula.
\begin{equation}
{\rm{AP}}{{\rm{S}}_{\rm{R}}}({\varphi _{\rm{R}}}) = \sum\limits_{l = 1}^L {{{\left| {{{\hat \gamma }_l}} \right|}^2}} \delta ({\varphi _{\rm{R}}} - {\hat \varphi _{{\rm{R}},l}}),
\end{equation}
where $\delta ( \cdot )$  is the Dirac delta function.
The above five types of channel parameters obtained above will serve as labels and prediction targets for model training. 

\subsubsection{Panoramic Image Post-processing}
\begin{figure}[!t]
	\centering
	\includegraphics[width=.4\textwidth]{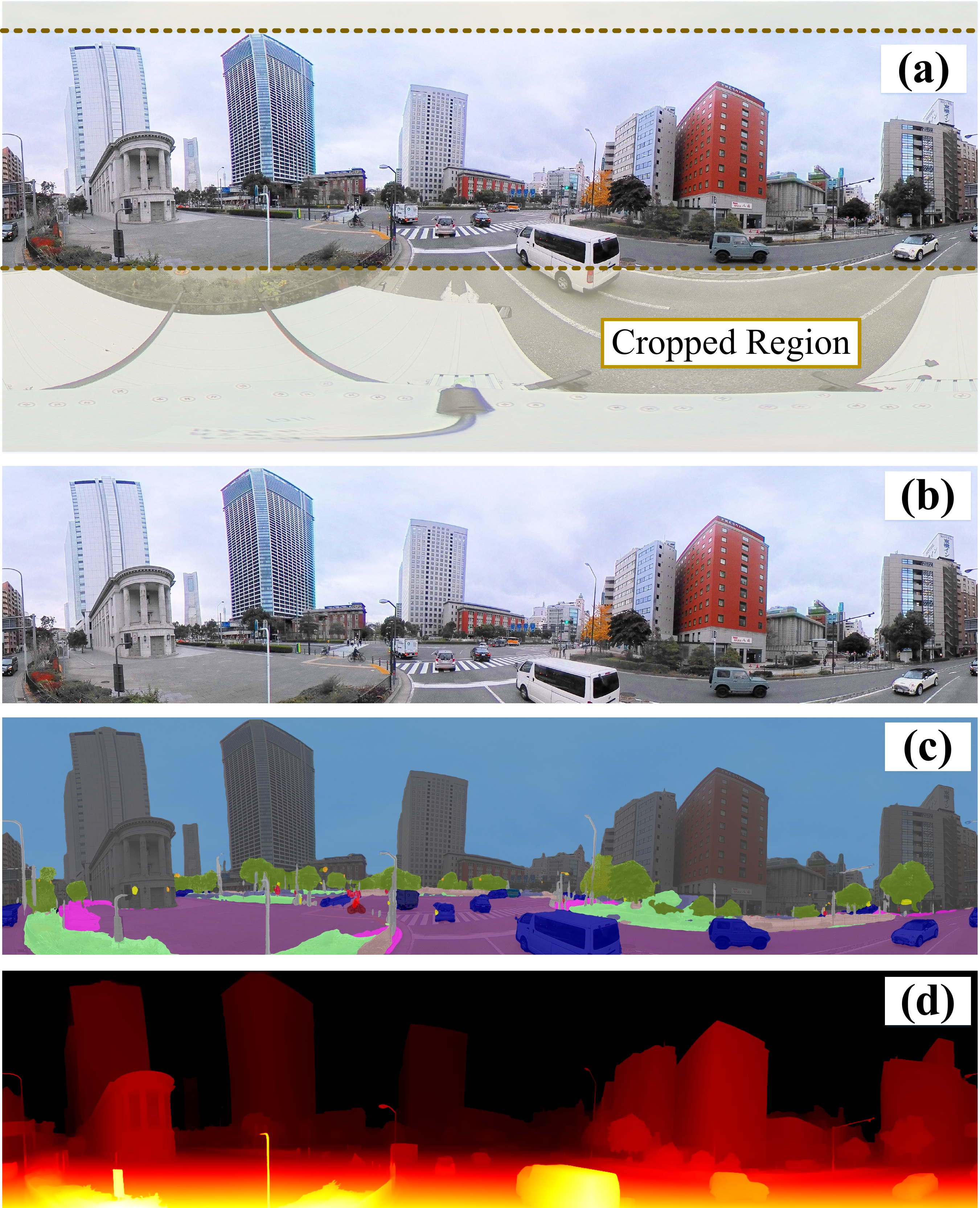}%
	\caption{Four representative types of panoramic images:
		(a) raw panoramic image captured by the rooftop-mounted camera, 
		(b) post-processed image after region cropping, 
		(c) semantic segmented  image, 
		(d) depth-estimated  image.
	}                    
	\label{image}
\end{figure}

The original panoramic images captured by the receiver-mounted camera, shown in Fig. \ref{image}(a), are stored in equirectangular projection with 360\degree\ azimuth and 180\degree\ elevation coverage. Since the vehicle body appears persistently in all frames and provides no useful information for training, each image is vertically cropped by about 75\degree\ to remove the vehicle region and by an additional 5\degree\ to exclude the sky, resulting in the processed panoramic images in Fig. \ref{image}(b) with full horizontal coverage and an effective vertical field of view of about 100\degree.

To provide the proposed model with propagation-relevant environmental representations, the cropped panoramic images are further processed into semantic and depth images. 
Specifically, \textit{Mask2Former} \cite{Cheng2022} and \textit{Depth Anything v2} \cite{Yang2024} are adopted for semantic segmentation and monocular depth estimation, respectively. Both models are initialized with pre-trained weights (trained on \textit{Cityscapes} \cite{Cordts2016}) and then fine-tuned on our dataset for domain adaptation. The resulting semantic and depth outputs are shown in Figs. \ref{image}(c) and (d), respectively. Since no absolute distance anchors are available during measurements, the generated depth images represent relative rather than metric depth; therefore, each depth image is normalized to $[0,1]$ before being used as model input.


\subsubsection{Multimodal Data Synchronization and Filtering}
After data post-processing, the multimodal data are further aligned in time and space to construct the final dataset. Since channel sounding signals, panoramic images, and GPS data are acquired at different rates, spatial consistency is ensured by the physical sensor configuration shown in Fig. \ref{system}, while temporal alignment is achieved through timestamp synchronization. As all three modalities are recorded with millisecond-level timestamps, samples with a time offset below 0.1 s are regarded as synchronized. In addition, invalid samples are removed, including those affected by low-SNR channel estimates, GPS drift, image occlusion, or prolonged vehicle stops, to ensure dataset reliability.

After synchronization and filtering, a multimodal dataset is obtained with a one-to-one correspondence among  panoramic images, channel data, and GPS records. 
Each sample is valid and non-duplicated, and no two samples are exactly identical.

\subsection{Model Training}
After   data post-processing, the channel-image joint dataset is thus obtained, i.e., $D = \left\{ (\Omega_m, \mathbf{Y}_m^\star) \right\}_{m=1}^M$.
As shown in Fig. \ref{framework}, semantic segmentation images, depth images, and GPS data are used as model inputs, while five channel parameters are used separately as training labels.
A total of 926 samples are collected, comprising 252 from \textit{Area 1}, 200 from \textit{Area 2}, 15 from \textit{Area 3}, and 459 from \textit{Area 4}.

Training hyperparameters are target-dependent. For PL/DS/ASA/ASD, the models in Fig. \ref{framework} are trained for 100 epochs with a batch size of 16 using Adam and MSE loss, with a learning rate of $10^{-3}$, weight decay of $10^{-6}$, learning-rate decay factor of 0.5, gradient clipping of 1, and early stopping with a patience of 20 epochs. 
For APS, the model  is also trained for 100 epochs, but with a batch size of 8, weight decay of $10^{-4}$, and AdamW optimizer. A module-wise learning-rate scheme is adopted, with ${3.5\times10^{-5}}$ for the semantic branch and ${3.5\times10^{-4}}$ for the remaining branches, together with a cosine-annealing warm-restart scheduler. Gradient clipping is set to 1.5, the loss in Eq. (\ref{math9}) is used, and early stopping follows the same 20-epoch patience. The detailed settings are summarized in Table \ref{hyperparam}.

\begin{table}[t!]
	\belowrulesep=0pt
	\aboverulesep=0pt
	\renewcommand{\arraystretch}{1.2}
	\centering
	\caption{Summary of Training Hyperparameter}
		\begin{tabular}{c|c|c}
			\toprule
			Hyperparameter & \makecell[c]{APS \\ Prediction Model} &  \makecell[c]{PL/DS/ASA/ASD \\ Prediction Model} \\
			\midrule
			Image Size & 224 & 224 \\
			Batch Size & 8 & 16 \\
			Max Epochs & 100 & 100 \\
			Weight Decay & $10^{-4}$ & $10^{-6}$ \\
			Gradient Clipping  & 1.5 & 1.0 \\
			Early Stopping & 20 Epochs& 20 Epochs \\
			Optimizer & AdamW & Adam \\
			Learning Rate  & \makecell[c]{$3.5 \times 10^{-5}$ / 
			$3.5 \times 10^{-4}$ }& $10^{-3}$ \\
			Loss Function & \makecell[c]{Multi-Constraint Loss }& MSE \\
			\bottomrule
		\end{tabular}
	\label{hyperparam}
\end{table}

\section{Performance  Evaluation and Analysis}
Based on the constructed  dataset, four validation experiments are conducted to evaluate the model performance: multi-parameter prediction under different modal inputs, dynamic scatterer removal analysis, backbone network impact and complexity analysis, and APS prediction validation. For convenience, they are denoted as Exp. 1 to Exp. 4, respectively.
Three evaluation metrics are selected to  evaluate the performance: root mean square error (RMSE), mean absolute error (MAE),
and cosine similarity, which primarily used to evaluate   APS prediction.

RMSE penalizes large errors and is therefore sensitive to outliers, reflecting the average deviation between predictions and ground truth, defined as
\begin{equation}
	\label{math12}
	\text{RMSE} = \sqrt {\frac{1}{M}\sum\limits_{m = 1}^M {{{({y_m} - \hat{y}_m)}^2}} },
\end{equation}
where $\mathbf{y} = \left\{ {{y_1},...,{y_M}} \right\}$ and $\mathbf{\widehat y} = \left\{ {\widehat {{y_1}},...,\hat{y}_M} \right\}$ refer to the ground truth and predicted values, respectively.

MAE is less sensitive to outliers and quantifies the average absolute deviation between predictions and ground truth and is formulated as
\begin{equation}
	\text{MAE} = \frac{1}{M}\sum_{m=1}^{M}|y_m - \hat{y}_m|.
\end{equation}


Cosine similarity quantifies the directional agreement between two vectors and takes values in $[ - 1,1]$, where values closer to 1 indicate higher similarity. 
It is invariant to vector magnitude and is therefore suitable for assessing feature similarity and is calculated as
\begin{equation}
\text{CosSim} = \frac{\sum_{m=1}^{M}y_m{\hat{y}_m}}{\sqrt{\sum_{m=1}^{M}y_m^2} \cdot \sqrt{\sum_{m=1}^{M}{\hat{y}_m}^2}}.
\end{equation}

\begin{table*}
	\belowrulesep=0pt
	\aboverulesep=0pt
	\renewcommand{\arraystretch}{1.2}  
	\setlength{\tabcolsep}{4pt}  
	\begin{center}
		\caption{Summary of Results from Different Verification Experiments.}
		\label{summary_results}
		\begin{threeparttable}
			\begin{tabular}{c|c|c|cccc|cccc|c}
				\toprule
				\multirow{3}{*}[0.25ex]{\textbf{Exps.}} &
				\multirow{3}{*}[0.25ex]{\makecell[c]{\textbf{Semantic} \\ \textbf{Branch} \\ \textbf{Backbone}}}   & \multirow{3}{*}[0.25ex]{\textbf{Input Modal\tnote{(1)}}} &
				\multicolumn{9}{c}{\textbf{Evaluation Metrics\tnote{(2)}}} \\
				\cline{4-12}
				& & & \multicolumn{4}{c|}{\textbf{RMSE}} & \multicolumn{4}{c|}{\textbf{MAE}} & \textbf{CosSim} \\
				\cline{4-12}
				& & & \textbf{PL [dB]} & \textbf{DS [ns]} & \textbf{ASA [$^\circ$]} & \textbf{ASD [$^\circ$]} & \textbf{PL [dB]} & \textbf{DS [ns]} & \textbf{ASA [$^\circ$]} & \textbf{ASD [$^\circ$]} & \textbf{APS} \\
				\midrule
				\multirow{4}{*}[0.3ex]{Exp. 1}   & \multirow{4}{*}[0.3ex]{ResNet-34}
				& M1 \& M2 \& M3 & 3.26 & 37.66 & 5.05 & 5.08 & 2.08 & 20.36 & 2.99 & 3.23 & --   \\
				& & M1 \& M2     & 5.31 & 40.30 & 5.26 & 5.36 & 3.12 & 25.33 & 3.36 & 3.68 & --   \\
				& & M1 \& M3     & 4.11 & 45.09 & 6.71 & 5.71 & 2.89 & 28.03 & 5.11 & 3.58 & --   \\
				& & M1  & 6.11   & 40.24 & 5.48 & 5.77 & 4.36 & 24.10 & 3.64 & 3.52 & --   \\
				\hline
				\multirow{2}{*}{Exp. 2}   & \multirow{2}{*}[0.3ex]{ResNet-34}
				& M1 \& M2 \& M3    & 4.98 & 46.40 & 6.98 & 4.02 & 2.87 & 27.71 & 4.83 & 2.72 & --   \\
				& & M1' \& M2' \& M3  & 6.59 & 89.64 & 5.71 & 3.42 & 5.02 & 72.87 & 3.47 & 1.99 & --   \\
				\hline
				\multirow{5}{*}[0.25ex]{Exp. 3}  & ResNet-34 
				& \multirow{5}{*}[0.25ex]{M1 \& M2 \& M3} 
				& 4.16 & 42.34 & 5.34 & 5.62 & 2.42 & 21.75 & 3.25 & 3.63 & --  \\
				& VGG-16      & & 12.77 & 148.13 & 19.94 & 20.01 & 10.24 & 100.07 & 14.81 & 16.02 & --   \\
				& AlexNet     & & 6.14  & 74.17 & 10.73 & 10.79 & 4.66  & 55.24 &  7.61 & 10.37 & --   \\
				& U-Net       & & 6.70  & 78.82 & 11.21 & 11.30 & 4.99  & 58.31 &  7.98 & 10.79 & --   \\
				& MobileNetV2 & & 12.56 & 145.79 & 19.59 & 19.68 & 10.26 & 98.85 & 14.69 & 15.87 & --   \\
				\hline
				Exp. 4   & ResNet-34
				& M1 \& M2 \& M3  & \multicolumn{4}{c|}{0.1866 (Normalized APS)} & \multicolumn{4}{c|}{0.1423 (Normalized APS)} & \makecell[c]{Mean: 0.934 \\ Median: 0.957 } \\   
				\hline
				\bottomrule
			\end{tabular}
			\vspace{0.4em}
			\parbox{\linewidth}{\footnotesize
				\textit{Notes}:  
				(1) M1: semantic  images,
				M2: depth images,
				M3: GPS data,
				M1': semantic  images with  dynamic scatterers removed,
				M2': depth images with  dynamic scatterers removed.
				(2) Cosine similarity is used only to evaluate APS, whereas RMSE and MAE are used  for PL/DS/ASA/ASD/APS.
			}
		\end{threeparttable}
	\end{center}
\end{table*}

\subsection{Multi-Parameter and Different Modal Inputs Verification}

\begin{figure}[!t]
	\centering
	\subfloat[]{\includegraphics[width=.45 \textwidth]{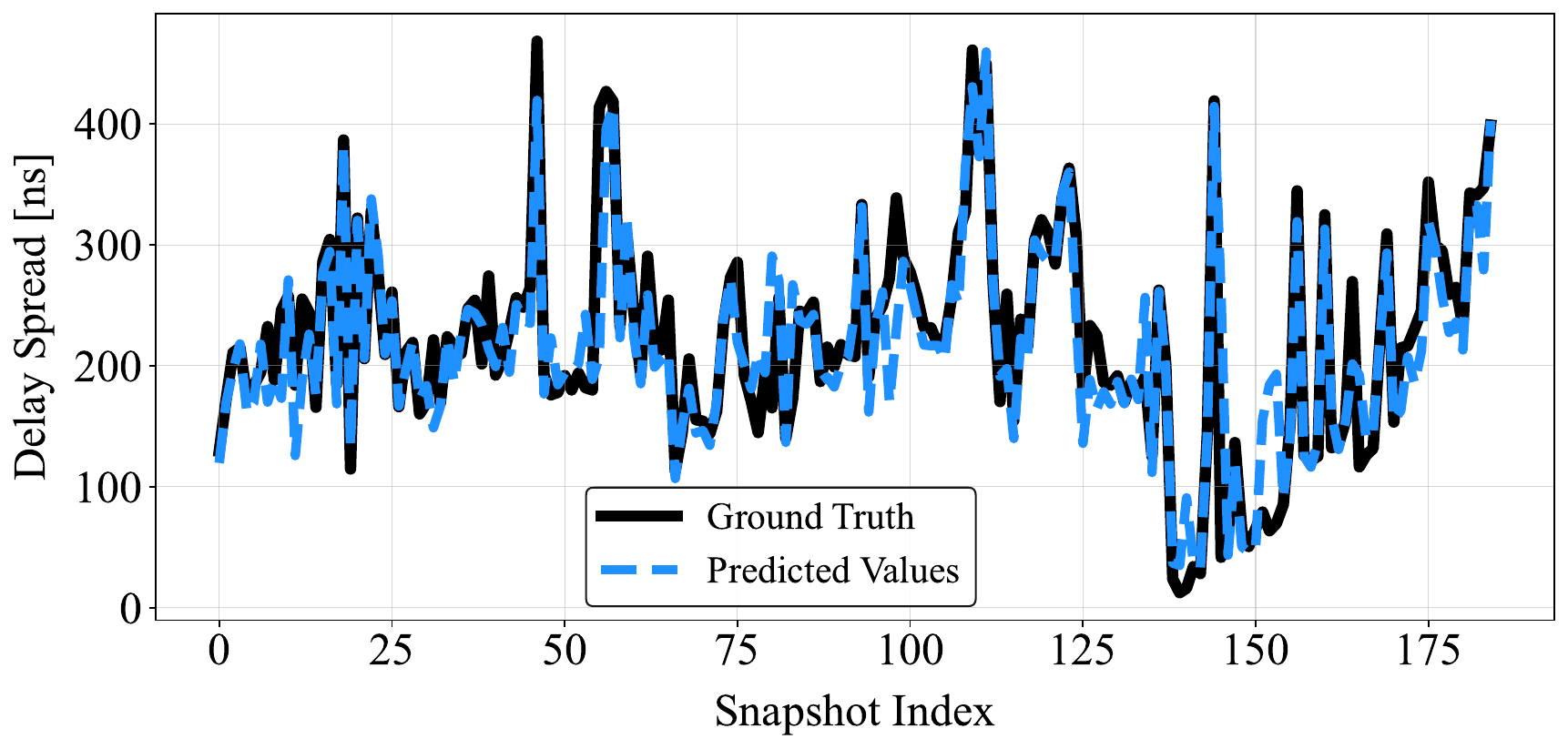}  
		\label{Exp1_DS}}
	\quad
	\subfloat[]{\includegraphics[width=.45 \textwidth]{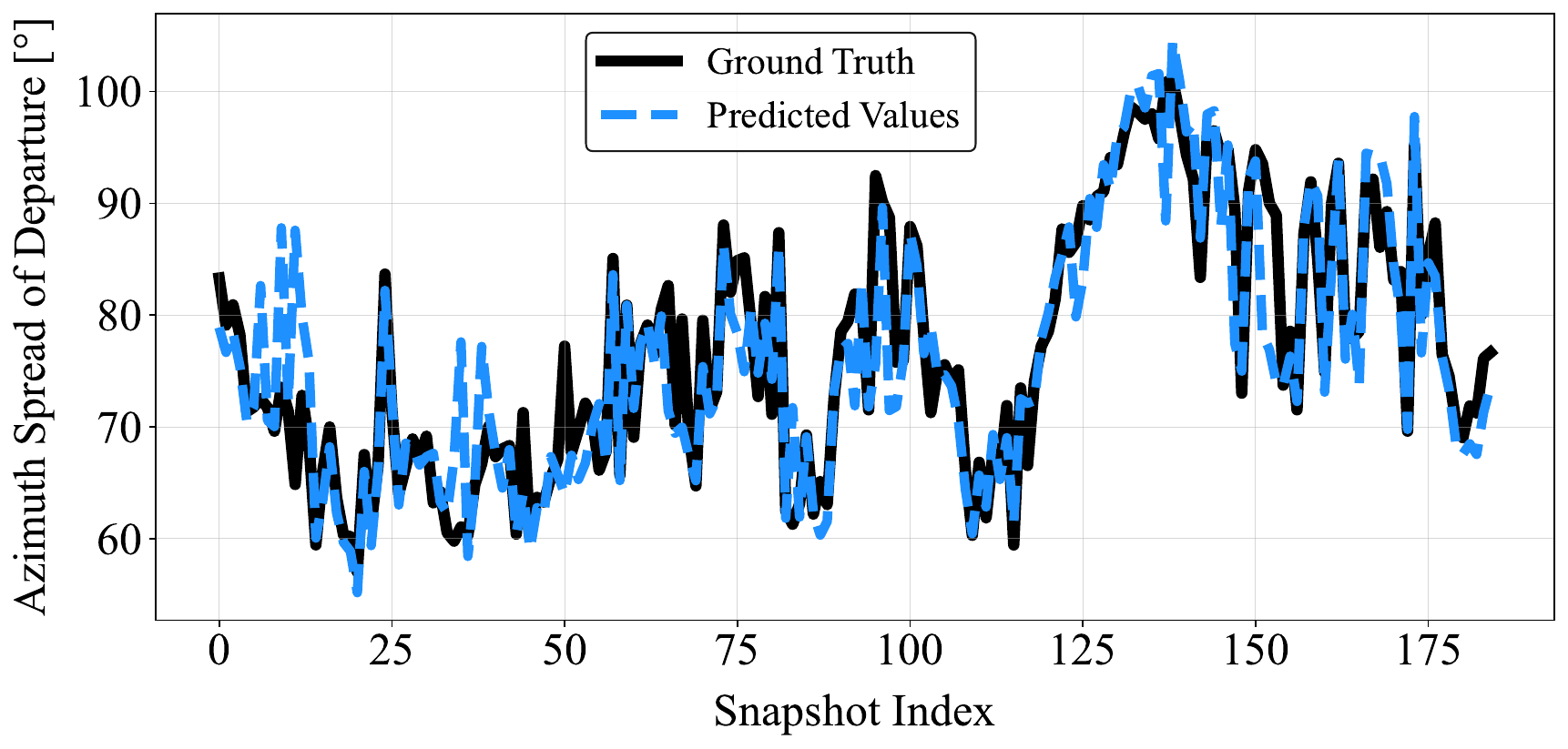}  
		\label{Exp1_ASD}}
	
	\caption{Prediction results of the proposed model for multiple large-scale parameters using semantic segmentation images, depth images, and GPS data inputs,
	taking (a) DS and (b) ASD as  examples.}
	\label{Exp1}
\end{figure}

\begin{figure}[!t]
	\centering
	\subfloat[]{\includegraphics[width=.24 \textwidth]{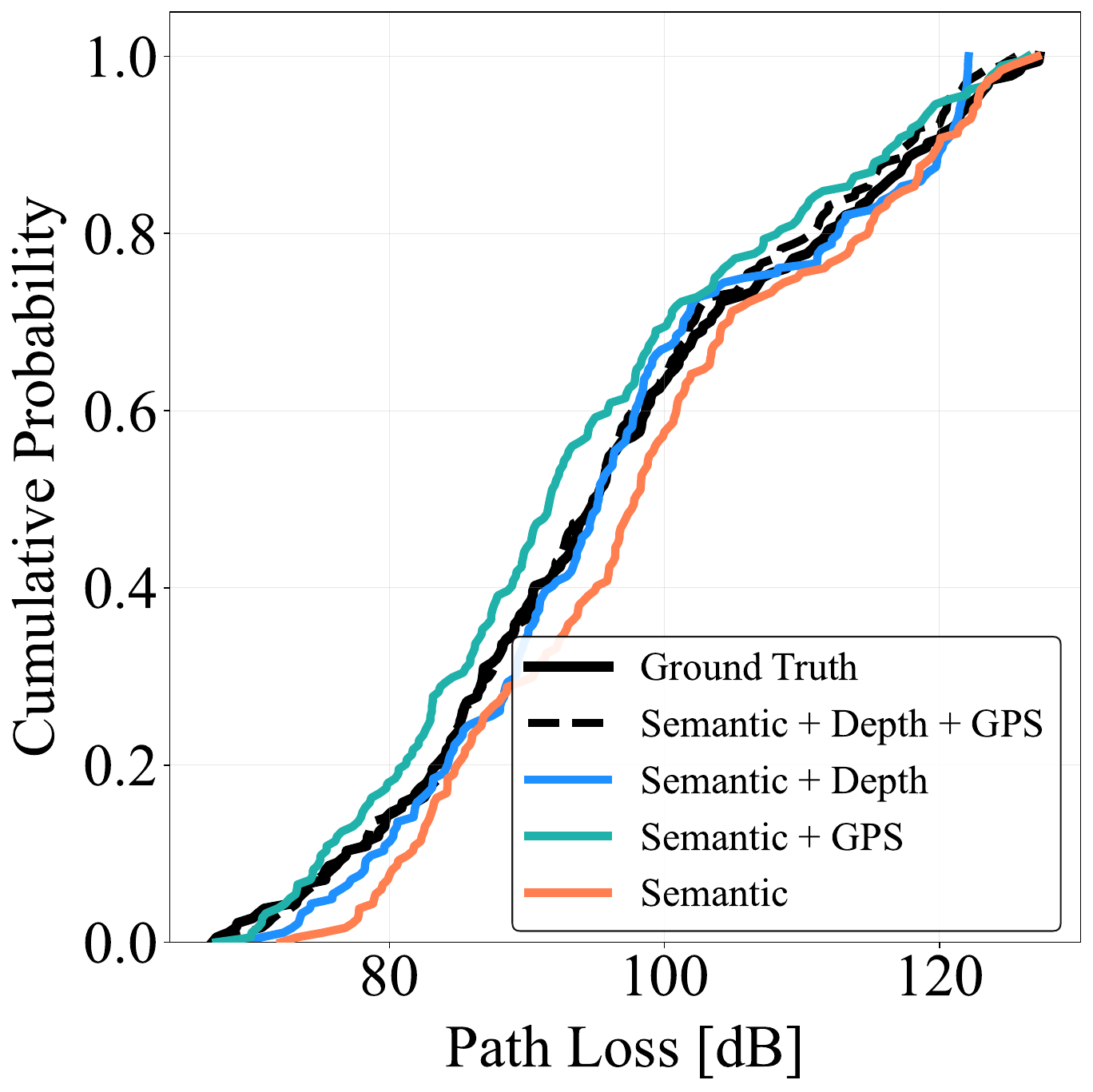}  
		\label{Exp2_PL}}
	\subfloat[]{\includegraphics[width=.24 \textwidth]{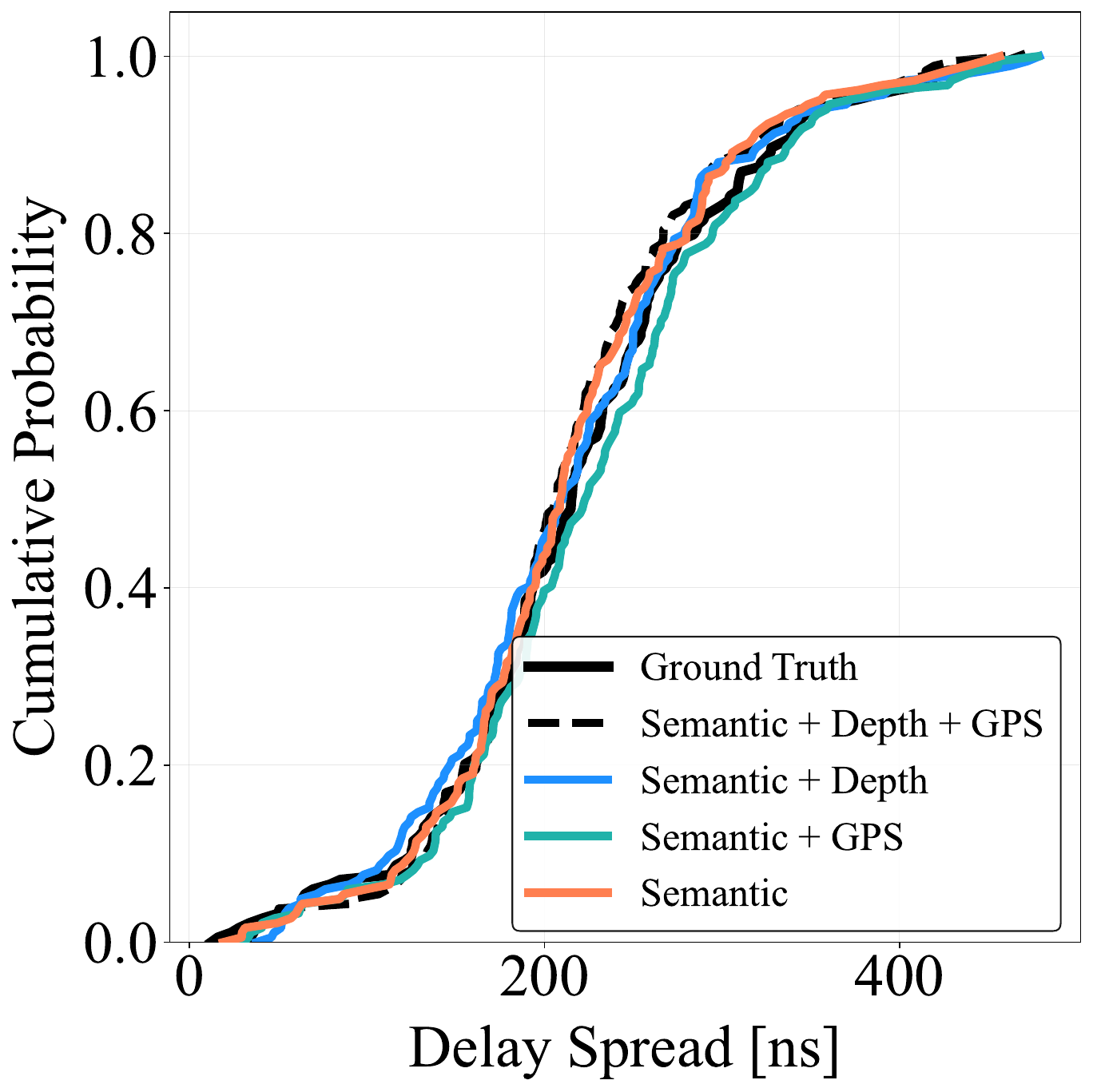}  
		\label{Exp2_DS}}
	\quad
	\subfloat[]{\includegraphics[width=.24 \textwidth]{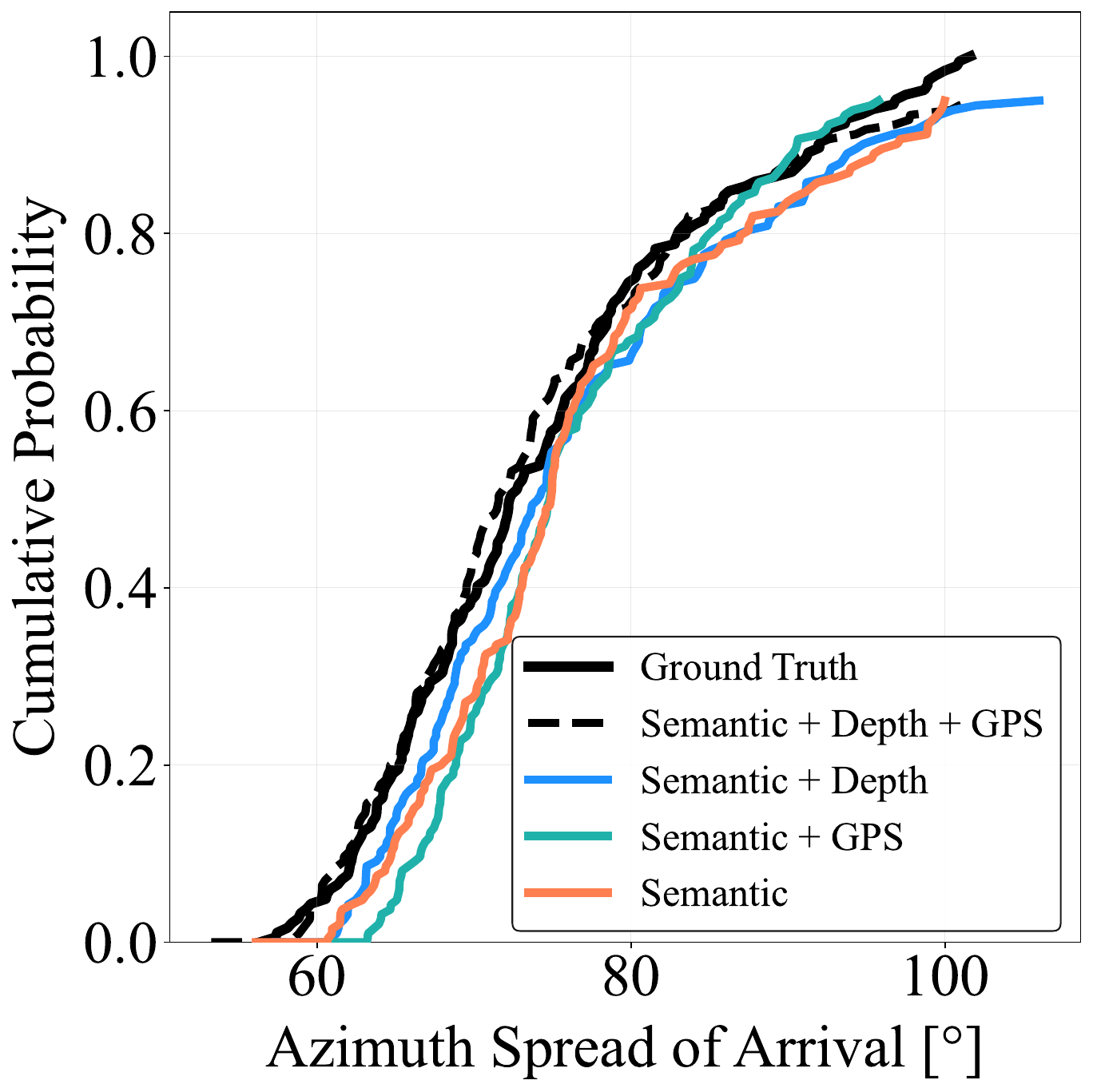}  
		\label{Exp2_ASA}}
	\subfloat[]{\includegraphics[width=.24 \textwidth]{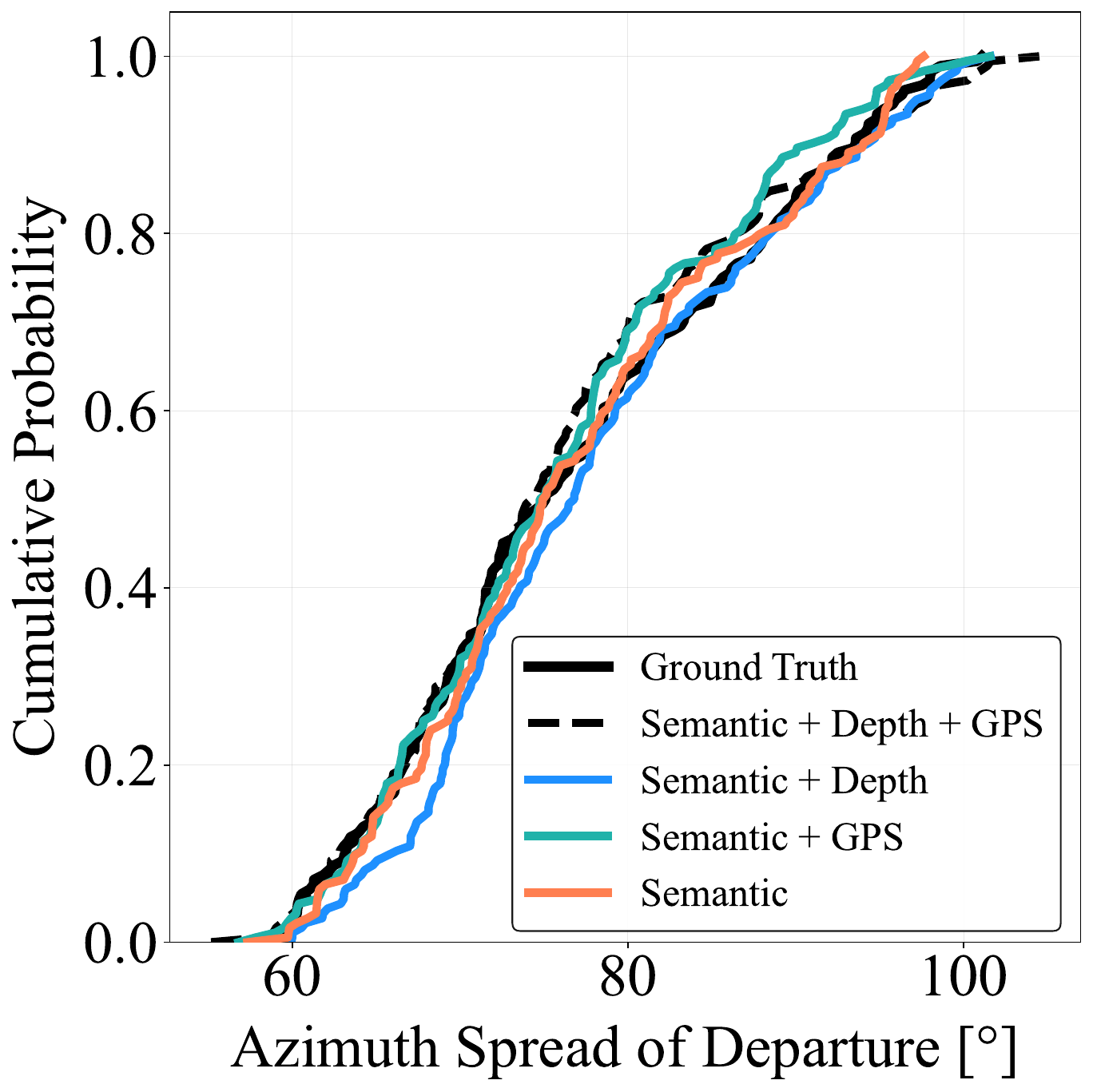}  
		\label{Exp2_ASD}}
	\caption{Results of the impact of different modal inputs on prediction accuracy:
		(a) PL;
		(b) DS;
		(c) ASA;
		(d) ASD.
	}
	\label{Exp2}
\end{figure}

This experiment evaluates the proposed framework for multiple large-scale channel prediction. Data from \textit{Areas 1-3} are used for training and validation, while a subset of \textit{Area 4} is reserved for testing. 
The prediction results  under the joint input of semantic  images, depth images, and GPS data are shown in Fig. \ref{Exp1}, and comparisons across different modality combinations are presented in Fig. \ref{Exp2}, with the corresponding RMSE and MAE summarized in Table \ref{summary_results}.

It can be observed that the  joint use of semantic  images, depth images, and GPS data achieves the best overall performance among all compared schemes, with predictions closely following the ground-truth trends. Under this setting, PL attains an RMSE/MAE of 3.26/2.08 dB, while the RMSEs of DS, ASA, and ASD are minimized at 37.66 ns, 5.05$^\circ$, and 5.08$^\circ$, respectively.
An ablation study is further conducted to quantify the contribution of each modality. When GPS data is excluded, the RMSE of PL  increases from 3.26 dB to 5.31 dB, accompanied by consistent performance degradation in DS, ASA, and ASD, indicating that GPS provides essential transceiver-level geometric priors. 
Removing depth images leads to a more pronounced decline in delay- and angle-domain prediction, with the RMSEs of DS and ASA increasing to 45.09 ns and 6.71$^\circ$, respectively, confirming the importance of depth cues for characterizing 3D geometry and occlusion. 
When only semantic segmentation images are used, the model exhibits the worst performance. The RMSE of PL  increases to 6.11 dB, and the remaining parameters also show the largest prediction errors. This suggests that semantic segmentation alone, although informative for environmental semantics such as scatterer categories and obstacle distribution, is insufficient for accurate multi-parameter prediction.

Overall, semantic segmentation images provide environment semantic  priors, depth images encode 3D propagation geometry, and GPS data introduces absolute transceiver-level positional constraints. 
These three modalities are highly complementary, and their joint use enables a more complete mapping between environmental features and channel characteristics, thereby achieving the most accurate prediction.



\subsection{Impact of Eliminating Dynamic Scatterers}

\begin{figure}[!t]
	\centering
	\includegraphics[width=.4\textwidth]{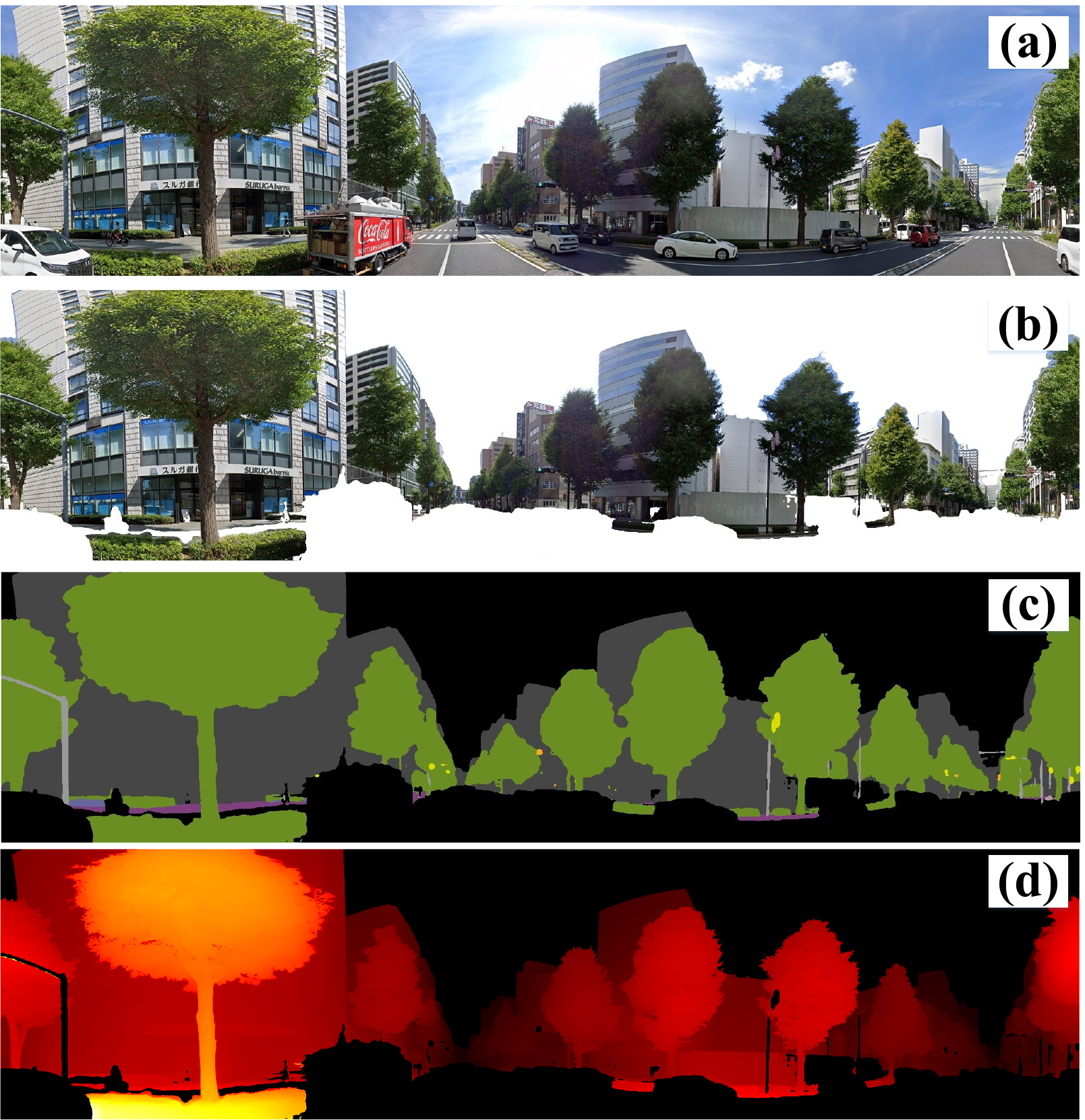}%
	\caption{Comparison of the raw image and  images with dynamic scatterers removed.
		(a) raw image;
		(b) raw image with dynamic scatterers removed;
		(c) semantic segmentation image with  dynamic scatterers  removed;
		(d) depth image  with  dynamic scatterers  removed. }                    
	\label{remove}
\end{figure}

In urban scenarios, large static scatterers, such as buildings, trees, and roads, primarily govern large-scale channel parameters, 
whereas small dynamic scatterers, such as vehicles and pedestrians, mainly induce local multipath fluctuations and instantaneous attenuation variations, and  have a relatively limited impact on channel statistics. 
Furthermore, the presence of dynamic scatterers substantially increases the complexity of visual data segmentation and annotation, requiring extensive manual annotation for accurate differentiation. 
This experiment therefore investigates whether dynamic scatterer removal significantly affects channel prediction performance.

\begin{figure}[!t]
	\centering
	\subfloat[]{\includegraphics[width=.45 \textwidth]{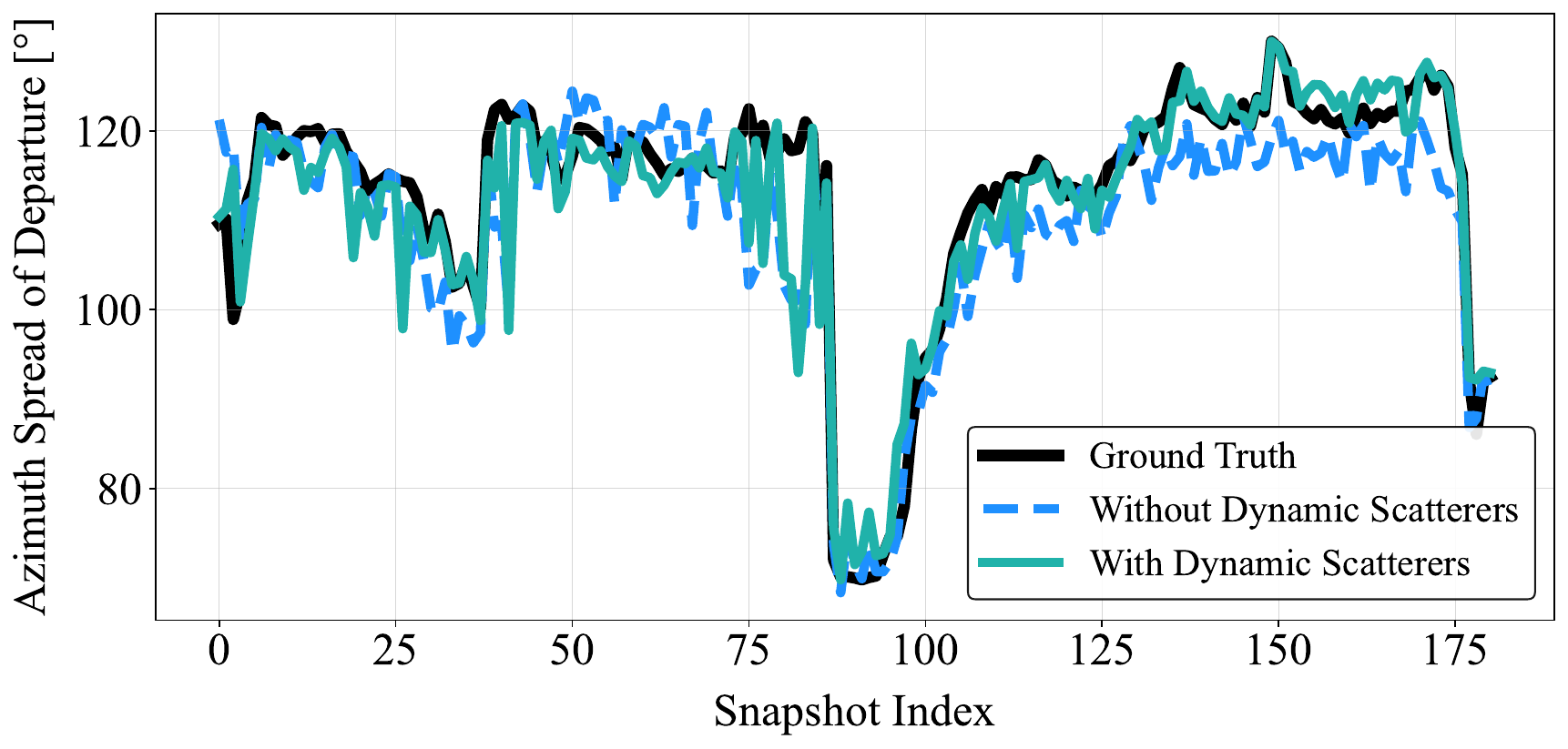}  
		\label{Exp3_PL}}
	\quad
	\subfloat[]{\includegraphics[width=.45 \textwidth]{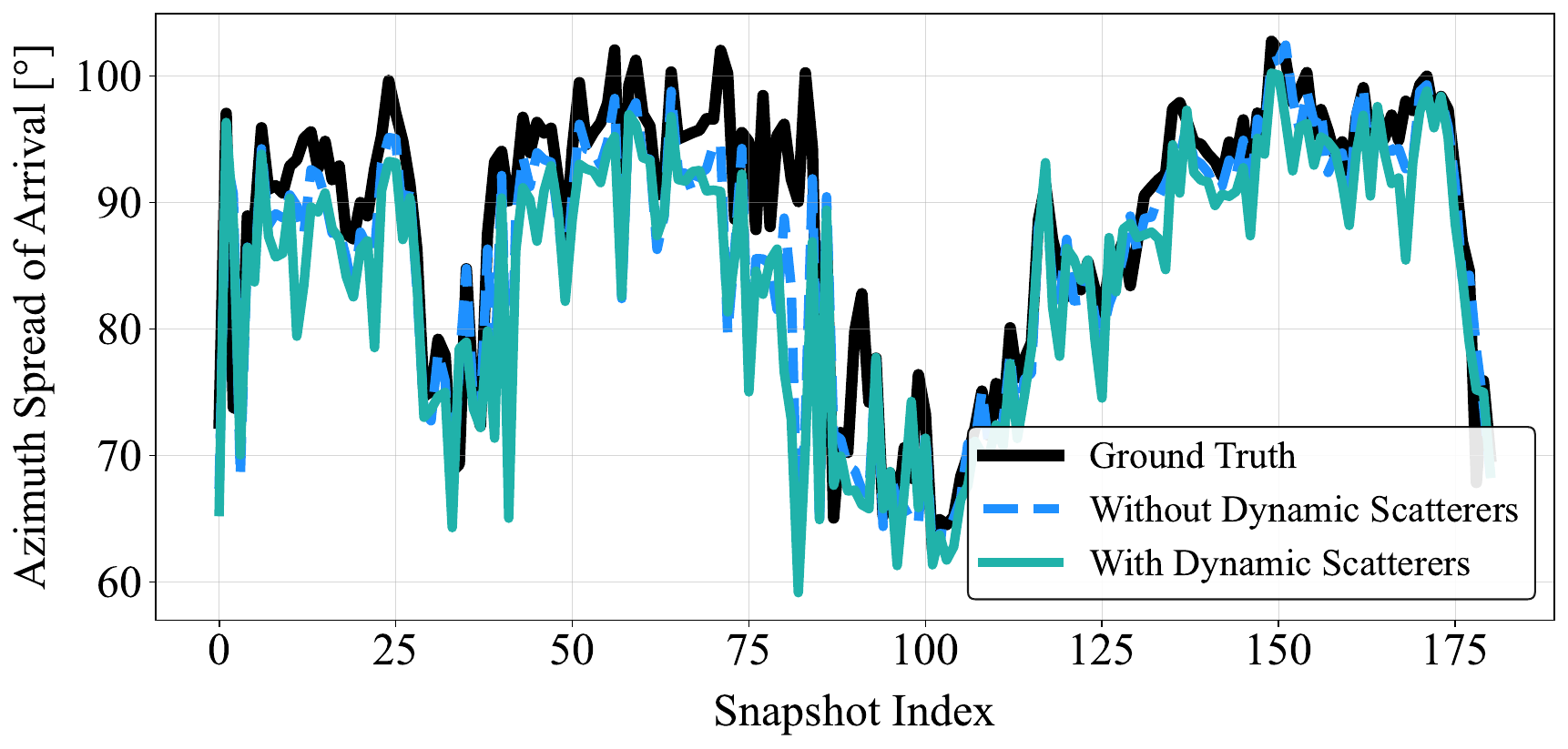}  
		\label{Exp3_ASD}}
	\caption{Results of the impact of eliminating dynamic scatterers on  prediction accuracy, taking (a) PL and (b) ASA as examples.
	}
	\label{Exp3}
\end{figure}

Following the post-processing procedure in Fig. \ref{dataset}, the dataset is re-annotated to generate multimodal images with dynamic scatterers removed, as shown in Fig. \ref{remove}. 
In the processed images, vehicles, pedestrians, the sky, and roads are masked for annotation simplicity. Correspondingly, these regions are represented in black in both semantic  images and depth images, with the masked regions in   depth images assigned the farthest distance.
Data from \textit{Area 1} is used as test set, while all data from \textit{Areas 2-4} are used for training. 
Since trimodal input achieves the best performance in previous experiments, this study only varies whether dynamic scatterer removal is applied, while consistently using semantic  images, depth images, and GPS data as input. 
The comparative results are shown in Fig. \ref{Exp3}, and the corresponding RMSE and MAE values are listed in Table \ref{summary_results}.

\begin{figure}[!t]
	\centering
	\includegraphics[width=.45\textwidth]{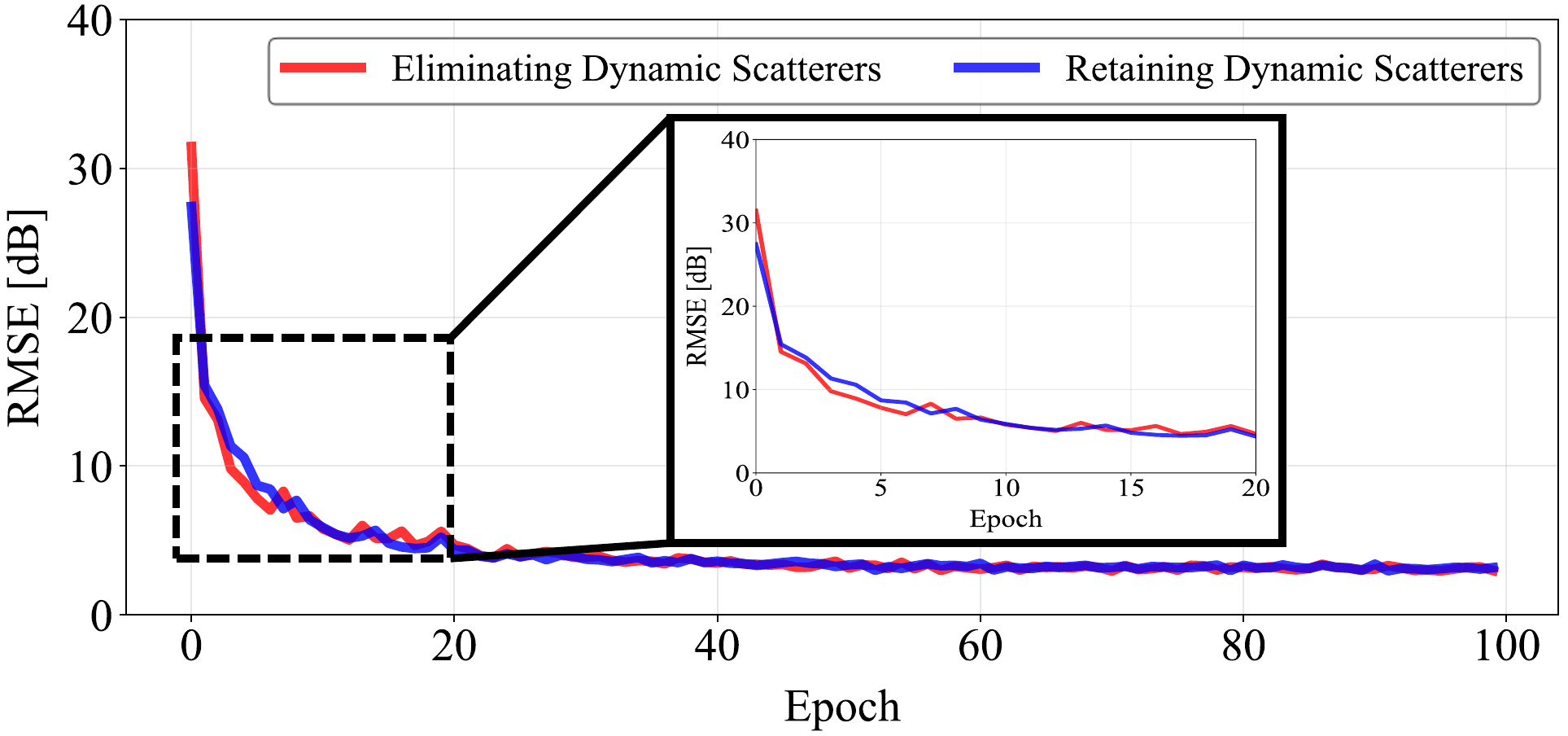}%
	\caption{Comparison of training error versus epoch with and without dynamic scatterer removal.}                    
	\label{Exp3_train}
\end{figure}

For PL and DS, dynamic scatterer removal significantly degrades prediction accuracy. 
The RMSE of PL increases from 4.98 dB to 6.59 dB, and that of DS rises from 46.40 ns to 89.64 ns. 
Although dynamic scatterers do not dominate the overall propagation trend, they remain part of the real environment and provide fine-grained propagation details through weak reflection, scattering, and additional attenuation.
Removing them causes the learned visual features to deviate from the actual scene, thereby reducing the fidelity of path loss and delay spread prediction.
By contrast, dynamic scatterer removal improves the prediction accuracy of angle-domain parameters. 
The RMSE of ASA decreases from 6.98$^\circ$ to 5.71$^\circ$, while that of ASD decreases from 4.02$^\circ$ to 3.42$^\circ$. 
This indicates that dynamic scatterers act mainly as irregular visual interference for angle estimation. Once removed, the model can focus more effectively on the dominant angular reflection and diffraction patterns determined by static scatterers, thereby improving angular prediction accuracy.
Training performance exhibits a similar trend. 
As shown in Fig. \ref{Exp3_train}, the model with dynamic scatterer removal converges faster and maintains lower training error across epochs. 
This is because removing dynamic scatterers suppresses random visual noise and allows the model to focus on static scatterer features that are more strongly correlated with the large-scale channel parameters.

In summary, dynamic scatterer removal improves training efficiency and benefits angle-domain parameter prediction, but degrades the prediction of path loss and delay spread by discarding environmental details that remain relevant to real propagation. Therefore, whether to apply dynamic scatterer removal should be determined by the target prediction task and the relative importance of different channel parameters.

%
%

\subsection{Impact of Different Backbone Networks and Complexity Analysis}

\begin{figure}[!t]
	\centering
	\includegraphics[width=.42\textwidth]{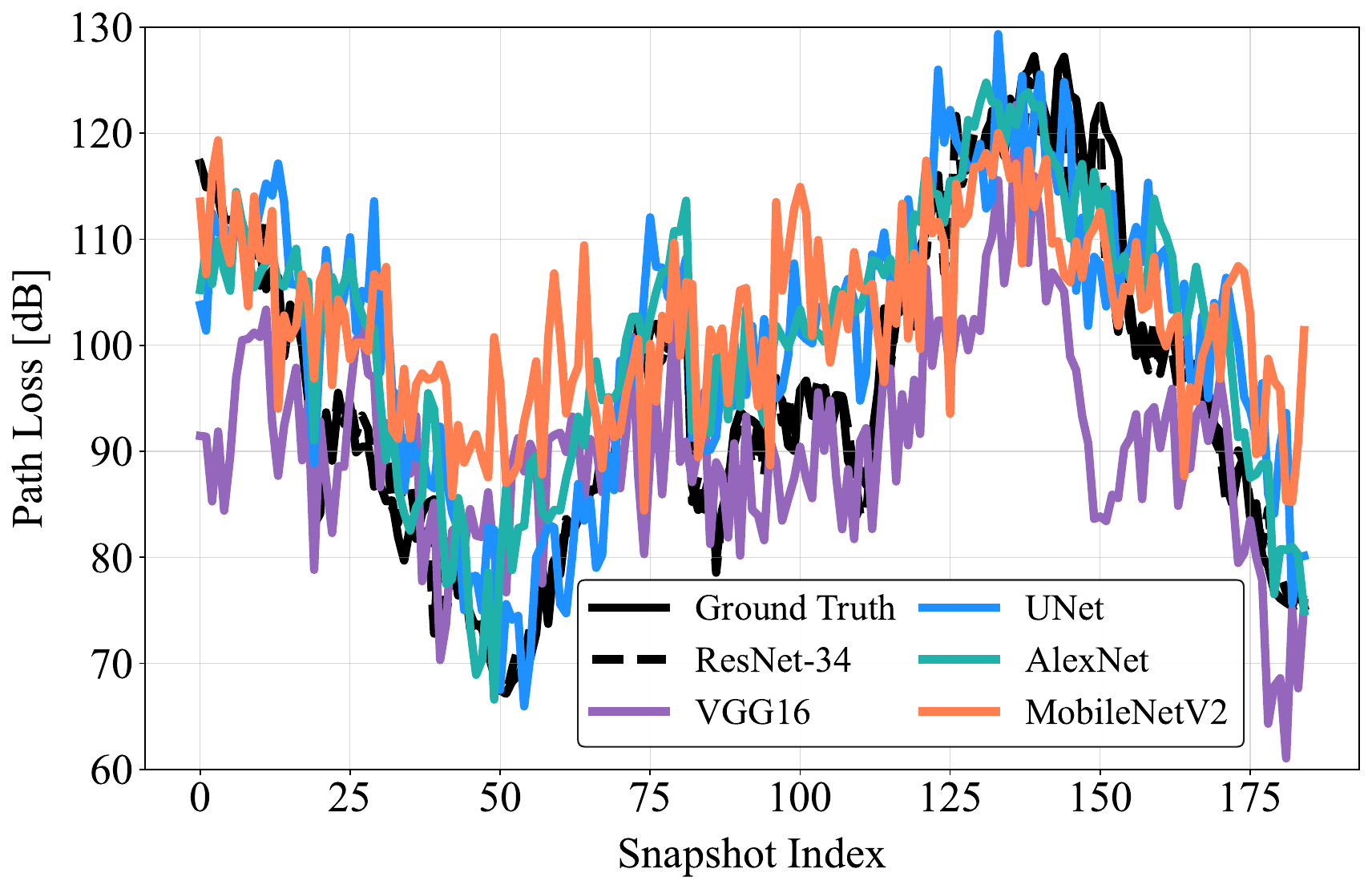}%
	\caption{Results of the impact of different backbone networks on prediction accuracy: taking PL as  the example. }                    
	\label{Exp5}
\end{figure}

This experiment maintains a fixed trimodal joint input configuration, replacing only the backbone network of semantic feature branch in Fig. \ref{framework}. 
The aim is to examine the impact of different backbones on the prediction performance and to analyze the relationship between predictive accuracy and model complexity. 
ResNet-34 is used as the baseline, with VGG-16, AlexNet, U-Net, and MobileNetV2 as comparison models. 
\textit{Area 2} is used for testing, while \textit{Area 1}, \textit{Area 3}, and \textit{Area 4} are used for training. 
The prediction performance comparison of PL is shown in Fig. \ref{Exp5}, and the corresponding error and complexity metrics are listed in Tables \ref{summary_results} and \ref{tab:model_complexity}.

It can be found that ResNet-34 achieves the best overall performance. 
Its RMSE and MAE for PL are 4.16 dB and 2.42 dB, respectively, and it is the only backbone that maintains high accuracy across PL, DS, ASA, and ASD. 
The other backbones perform notably worse. 
VGG-16 and MobileNetV2 yield the largest PL errors, with RMSE values of 12.77 dB and 12.56 dB, respectively, and fail to capture the variation trend effectively. 
AlexNet and U-Net perform moderately better, with PL RMSE values of 6.14 dB and 6.70 dB, respectively.
Although they can roughly predict the variation trend, they lag significantly behind ResNet-34 in fitting detailed fluctuations.

The complexity results show that predictive performance is not determined solely by parameter count, computational cost, or inference speed. 
Instead, it mainly depends on feature extraction capability and the suitability of  architecture for channel prediction task. 
ResNet-34 contains 20.82 M parameters and 7.37 G FLOPs, with 2.82 ms inference time and 709.82 FPS, indicating a favorable balance between representation capability and efficiency. U-Net has comparable complexity but is better suited to dense image prediction than to channel parameter regression. AlexNet is efficient but too shallow to learn sufficiently discriminative features. VGG-16 suffers from parameter redundancy and low efficiency. MobileNetV2 has the fewest parameters, but its lightweight design does not provide adequate representational capacity for this task.

Overall, the effectiveness of semantic branch backbone depends on both feature extraction capability and task matched complexity. 
ResNet 34 offers the best balance and is therefore the most suitable backbone. 
These results indicate that vision based channel prediction should prioritize strong feature representation and architectural suitability, rather than simply pursuing lightweight design or higher computational cost.

\begin{table}
	\belowrulesep=0pt
	\aboverulesep=0pt
	\renewcommand{\arraystretch}{1.2}  
	\setlength{\tabcolsep}{3pt}   
	\begin{center}
	    \caption{Complexity Comparison of Different Semantic Backbones}
        \label{tab:model_complexity}
		\begin{threeparttable}
 	     	\begin{tabular}{c|c|c|c|c}
				\toprule  
				\textbf{Model}      & \makecell[c]{\textbf{Total Params } \\ \textbf{(M)}}  & \makecell[c]{\textbf{FLOPs} \\ \textbf{(G)}} & \makecell[c]{\textbf{Avg Inference Time} \\ \textbf{  (ms)}}  &  \makecell[c]{\textbf{Inference} \\ \textbf{FPS}} \\
				\midrule  
				ResNet-34     & 20.82       & 7.37     & 2.82  &  709.82  \\
				VGG-16         & 120.94      & 30.93    & 3.25   & 616.3 \\
				AlexNet       & 47.26       & 1.42     & 0.9   & 2222.25\\
				U-Net          & 18.63       & 26.18    & 2.71   &  739.34   \\
				MobileNetV2   & 2.83        & 664.78     & 3.09    & 647.25  \\
				\bottomrule  
			\end{tabular}
    \vspace{0.4em}
    \parbox{\linewidth}{\footnotesize
    	\textit{Notes}: All experiments are conducted on an NVIDIA RTX 5080 GPU with batch size = 2 and input resolution = 224×224. ``FLOPs" denotes floating-point operations (G = giga), ``M" denotes million, and ``Inference FPS" denotes  the number of samples processed per second.}
	   \end{threeparttable}
  \end{center}
\end{table}

\subsection{APS Predictive Performance Verification}
\begin{figure}[!t]
	\centering
	\subfloat[]{\includegraphics[width=.24 \textwidth]{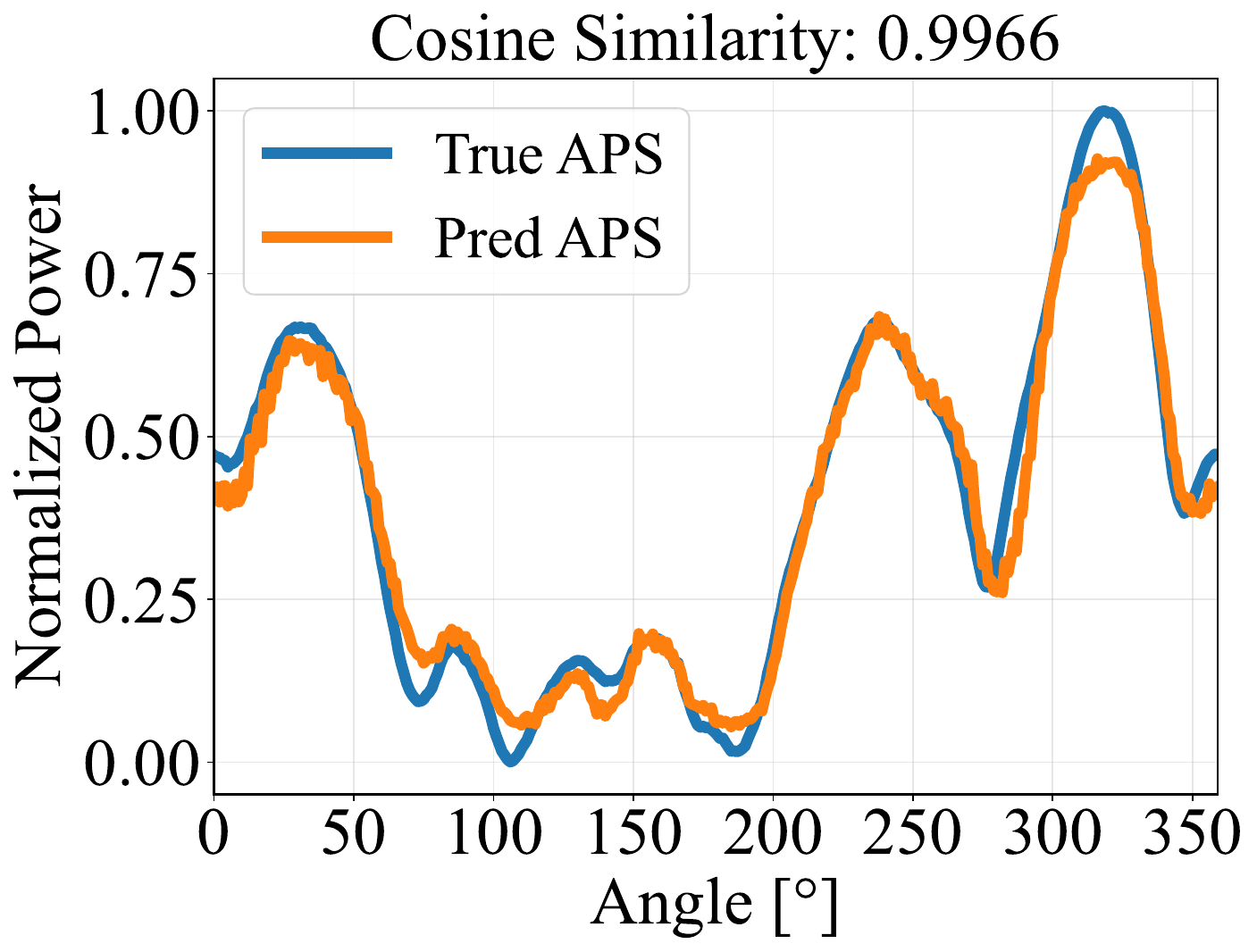}  
		\label{APS2}}
	\subfloat[]{\includegraphics[width=.24 \textwidth]{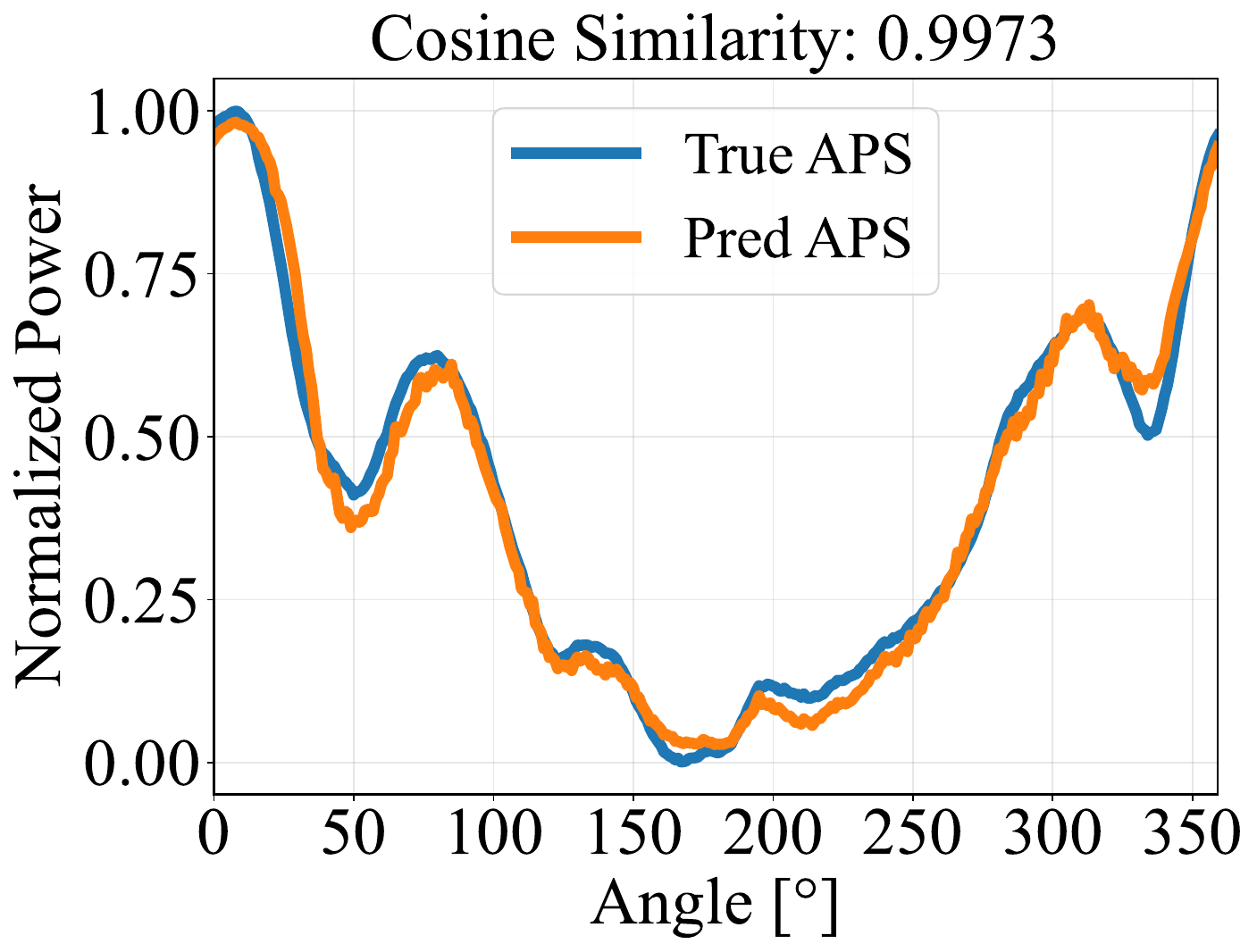}  
		\label{APS3}}
	\quad
	\subfloat[]{\includegraphics[width=.24 \textwidth]{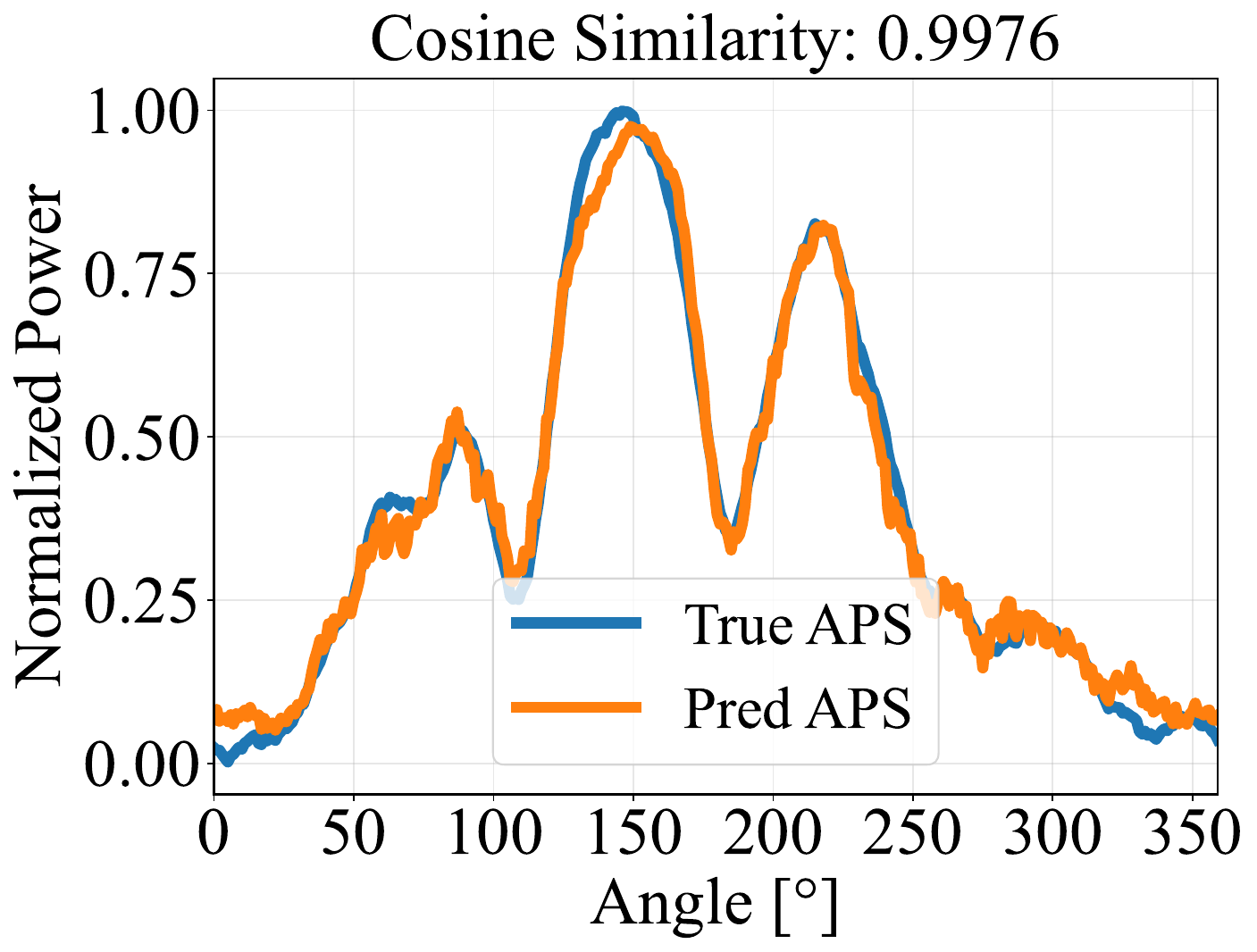}  
		\label{APS5}}
	\subfloat[]{\includegraphics[width=.24 \textwidth]{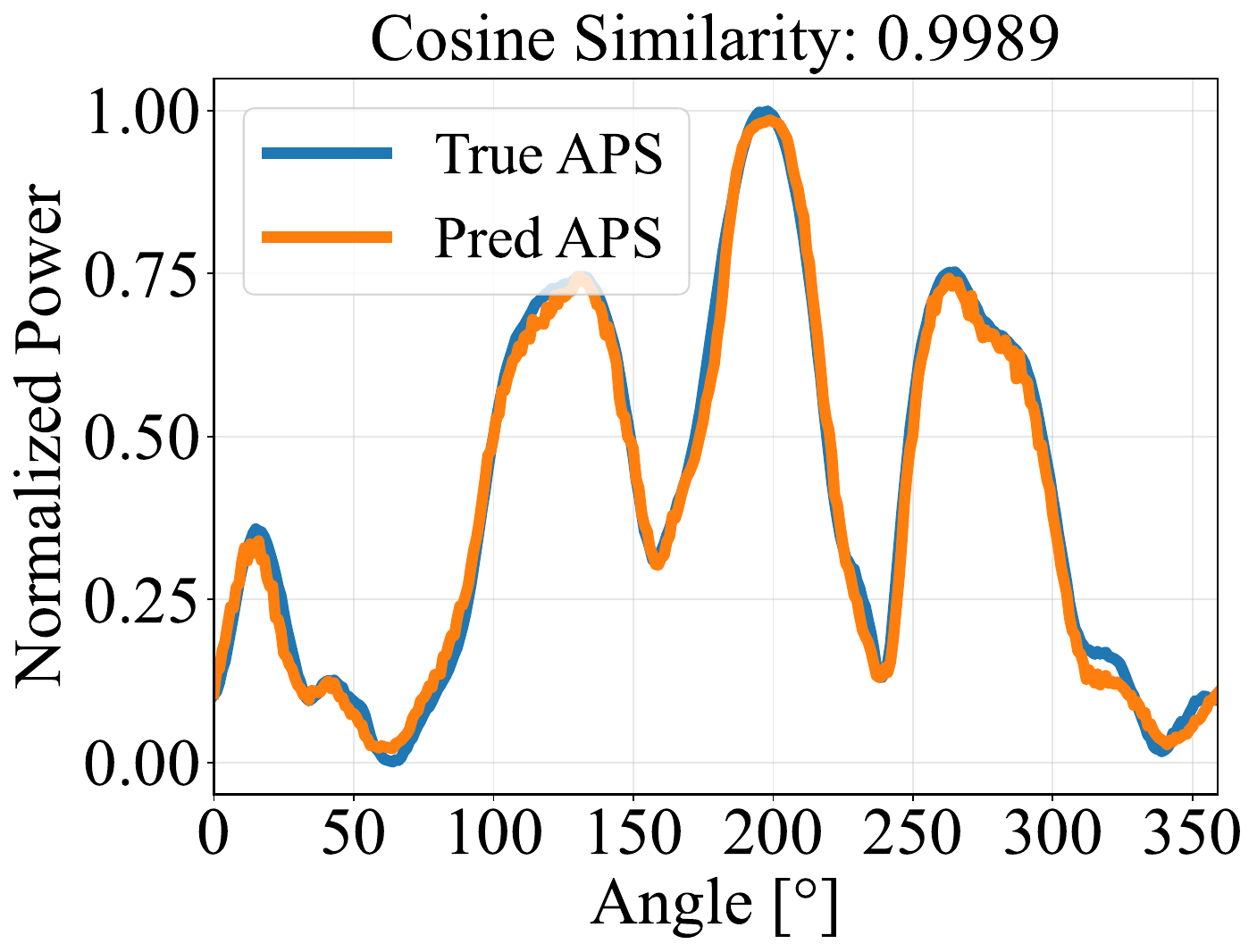}  
		\label{APS6}}
	\caption{Examples of APS prediction results.
	}
	\label{APS}
\end{figure}

\begin{figure}[!t]
	\centering
	\includegraphics[width=.42\textwidth]{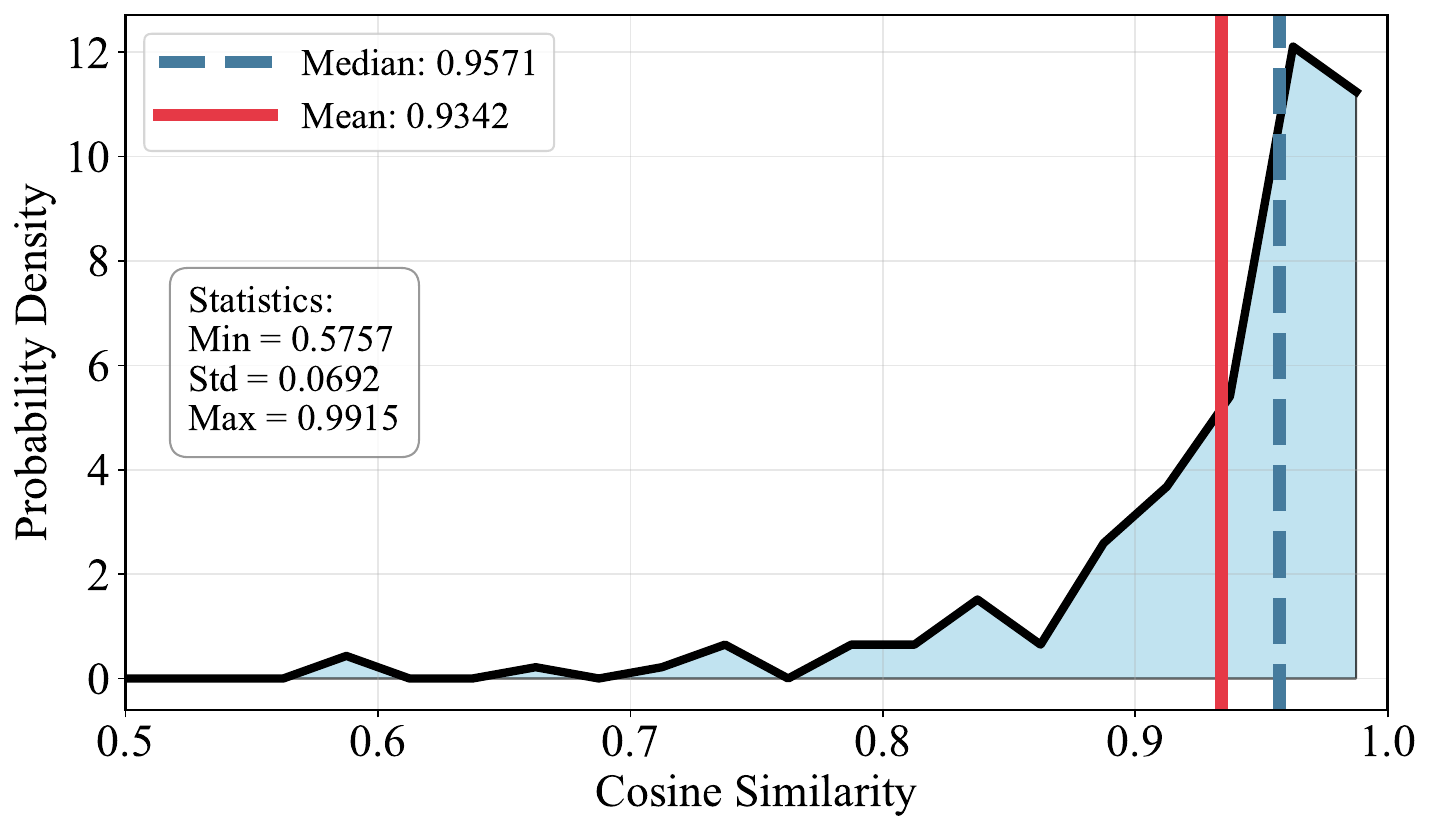}%
	\caption{Probability density distribution of cosine similarity of all samples in the test set. }                    
	\label{Exp6}
\end{figure}

APS  is a 360-dimensional vector that describes the angular distribution of multipath energy. 
Its prediction accuracy therefore directly reflects the  ability of model to capture fine grained angular domain structure. 
The experiment still takes   semantic segmentation image, depth image, and GPS data as trimodal input. 
Data from \textit{Area 4} is used for testing, while that from \textit{Areas 1-3} are used for training. 
The quantitative results are reported in Table \ref{summary_results}, representative snapshots are shown in Fig. \ref{APS}, and the cosine similarity distribution over  test set is given in Fig. \ref{Exp6}.

It can be found that the proposed model achieves excellent performance in normalized APS prediction. 
The RMSE and MAE are 0.1866 and 0.1423, respectively, while the mean and median cosine similarities reach 0.9342 and 0.9571. Over the full test set,  cosine similarity ranges from 0.5757 to 0.9915, with a standard deviation of 0.0692, and the probability density is concentrated above 0.9. These results indicate that the predicted APS is highly consistent with the ground truth for most samples, demonstrating good stability and robustness.

Fig. \ref{APS} presents   APS  prediction results for four snapshots with different main lobe locations and multipath distributions, which clearly demonstrate the  ability to capture fine angular domain structures. 
Taking Fig. \ref{APS}(a) as an example, the true APS exhibits a dominant main lobe in the 300$^\circ$ to 350$^\circ$ range, together with distinct low energy sidelobe fluctuations near 50$^\circ$ and 150$^\circ$. 
The predicted APS not only accurately identifies the main lobe location and peak amplitude, but also closely reproduces the sidelobe decay trend and local fluctuation details. 
Cosine similarity reaches 0.9966. 
In Fig. \ref{APS}(c), the main lobe is centered in the 100$^\circ$ to 150$^\circ$ range and shows a more complex pattern with one dominant lobe and multiple sidelobes. 
The predicted curve accurately matches the main lobe width and peak value, while also preserving the sidelobe attenuation behavior beyond 200$^\circ$, yielding a cosine similarity of 0.9976. Similar behavior is observed in Figs. \ref{APS}(b) and (d), where cosine similarities are 0.9973 and 0.9989, respectively. 
These results indicate that, regardless of the main lobe position, the model consistently captures both the dominant structure and subtle details of  APS, with the predicted curves closely overlapping ground truth.

In summary, the proposed multimodal fusion architecture efficiently models the angular domain energy distribution characteristics of  APS. 
The results confirm that this architecture can balance the overall stability and local precision of APS prediction, providing crucial support for achieving comprehensive coverage from scalar statistical features to vector-level angular domain channel representation in 6G wireless propagation models, fully validating its versatility and effectiveness in complex channel  prediction tasks.

\section{CONCLUSION} 

This paper proposes a deep learning framework for environment-aware channel prediction based on multimodal visual feature fusion for vehicular communications. 
By jointly exploiting vehicle-side panoramic visual data and positional information, the proposed framework unifies semantic, depth, and location modalities within a three-branch architecture with SE-guided adaptive fusion, enabling the joint prediction of PL, DS, ASA, ASD, and APS. 
A dedicated regression head and a composite multi-constraint loss are further designed for high-dimensional APS prediction.
Experimental results on a synchronized urban V2I measurement dataset demonstrate the effectiveness of the proposed framework.
Under trimodal input, the best performance is achieved, with an RMSE/MAE of 3.26/2.08 dB for PL, RMSEs of 37.66 ns, 5.05°, and 5.08° for DS, ASA, and ASD, respectively, and mean/median APS cosine similarities of 0.934/0.957. 
The results further suggest that dynamic-scatterer handling should be task-specific, since removing dynamic objects improves angular-domain prediction and convergence but degrades PL and DS estimation; 
meanwhile, ResNet-34 offers the best accuracy–efficiency trade-off.
Overall, this work verifies that multimodal environment-aware learning can effectively bridge visual sensing and  channel prediction in real V2I scenarios, offering a practical solution for intelligent channel prediction in future 6G vehicular communications.

	 \balance
	\bibliographystyle{IEEEtran}
	\nocite{*}
	
	\bibliography{IEEEabrv,ref}

\end{document}